\documentclass[letterpaper]{article} 
\usepackage{aaai2027}  
\usepackage[hyphens]{url}  
\usepackage{graphicx} 
\usepackage{natbib}  
\usepackage{caption} 
\usepackage{algorithm}
\usepackage{algorithmic}

\usepackage{newfloat}
\usepackage{listings}
\DeclareCaptionStyle{ruled}{labelfont=normalfont,labelsep=colon,strut=off} 
\floatstyle{ruled}
\newfloat{listing}{tb}{lst}{}
\floatname{listing}{Listing}

\usepackage{booktabs}

\usepackage{multirow} 
\usepackage{booktabs} 
\usepackage{multirow}
\usepackage{amsmath}
\usepackage{graphicx}   
\usepackage{multirow}   
\usepackage{array}      
\usepackage{booktabs}
\usepackage{multirow}
\usepackage{makecell}
\usepackage{graphicx}
\usepackage{tabularx}
\usepackage{multirow}    
\usepackage{booktabs}    
\usepackage{graphicx} 
\usepackage{booktabs}    
\usepackage{multirow}    
\usepackage{graphicx}    
\usepackage{amsmath}
\usepackage{algorithm}
\usepackage{algorithmic}

\usepackage{newfloat}
\usepackage{listings}
\usepackage{amsmath} 
\usepackage{amssymb}
\usepackage{multirow}
\usepackage{times}

\usepackage{booktabs}
\usepackage{multirow}
\usepackage{xcolor}
\usepackage{makecell}
\usepackage{graphicx}
\usepackage{fontawesome5}

\usepackage{latexsym}
\usepackage{amsmath} 
\usepackage{amssymb}
\usepackage{booktabs}
\usepackage{xcolor}
\usepackage{multirow}
\usepackage{graphicx}
\usepackage{fontawesome5} 
\usepackage{booktabs}
\usepackage{xcolor}
\usepackage{graphicx}
\usepackage{amsmath}
\usepackage{booktabs}
\usepackage{multirow}
\usepackage[table]{xcolor}
\usepackage{graphicx}
 \usepackage{colortbl} 
 \usepackage[utf8]{inputenc}
\usepackage[T1]{fontenc}
\usepackage{tikz}
\usepackage[most]{tcolorbox}
\usepackage[most]{tcolorbox}
\usepackage{array}
\usepackage{xcolor}

\definecolor{highColor}{RGB}{0,102,204}    
\definecolor{midColor}{RGB}{102,153,0}     
\usepackage{tabularx}
\usepackage{listings}
\usepackage{xcolor}
\usepackage{enumitem}

\definecolor{highColor}{RGB}{0, 102, 204}    
\definecolor{midColor}{RGB}{0, 153, 76}     
\definecolor{lowColor}{RGB}{128, 128, 128}   
\definecolor{codeBg}{RGB}{245, 245, 245}

\newtcolorbox{HighSkill}[2]{
    enhanced,
    colback=white,
    colframe=highColor,
    fonttitle=\bfseries\large,
    title={#1 \quad \normalsize },  
    attach boxed title to top left={xshift=3mm, yshift=-2mm},
    boxed title style={colback=highColor},
    drop shadow,
    before skip=15pt,
    after skip=15pt,
}

\newtcolorbox{MidSkill}[1]{
    enhanced,
    boxrule=0pt,
    frame hidden,
    left=3mm,
    overlay={
        \fill[midColor] (frame.north west) rectangle (frame.south west);
    },
    colback=midColor!5!white,
    sharp corners,
    title=\textbf{#1},          
    fonttitle=\bfseries\color{midColor!80!black},  
    colbacktitle=midColor!5!white,  
    title style={
        anchor=north west,
        at={(frame.north west)},
        xshift=0pt, yshift=0pt
    },
    before skip=10pt,
}

\usepackage{array}    
\usepackage{calc}     
\usepackage{tabularx} 
\usepackage{graphicx} 
\newcolumntype{C}{>{\centering\arraybackslash}m{2cm}}

\usepackage[T1]{fontenc}

\usepackage[utf8]{inputenc}

\usepackage{microtype}

\usepackage{inconsolata}

\usepackage{graphicx}

\title{CodeSkill: Latent Skill Abstraction for Long-Horizon Code Agents}
\author{
    Song-Li Wu\textsuperscript{\rm 1}\equalcontrib,
    Jingyi Wang\textsuperscript{\rm 1}\equalcontrib,
    Zhaocheng Du\corresponding\textsuperscript{\rm 2},
    Weinan Gan\textsuperscript{\rm 2},
    Weiwen Liu\textsuperscript{\rm 3}
}
\affiliations{
    \textsuperscript{\rm 1}Tsinghua University,
    \textsuperscript{\rm 2}Huawei Noah’s Ark Lab, 
    \textsuperscript{\rm 3}Shanghai Jiao Tong University
}

\begin{document}

\maketitle

\begin{abstract}
Code agents require long-horizon decision-making over complex interaction trajectories. However, existing reinforcement learning (RL) approaches typically optimize behavior at the token level, creating a mismatch between low-level generation and high-level behavioral reasoning. This limitation leads to inefficient exploration and weak credit assignment under sparse rewards. Moreover, while large-scale agent trajectories often contain recurring multi-step behavioral patterns, their noisy token-level representations hinder effective experience reuse. To address these challenges, we propose CodeSkill, a framework that adapts hierarchical latent skill modeling to the code agent domain. CodeSkill first leverages a teacher model to distill both successful and failed trajectories into multi-level textual abstractions. It then integrates temporal variational inference with reinforcement learning to map these discrete semantics into continuous latent variables, while an adaptive boundary mechanism dynamically gates skill transitions based on execution feedback. The learned skills are injected into a frozen LLM policy as latent semantic prefixes, enabling optimization in a compact semantic space rather than over raw token sequences. By shifting RL from token-level exploration to experience-level reasoning, CodeSkill improves optimization efficiency and long-horizon behavioral coherence. Extensive experiments demonstrate that CodeSkill achieves highly competitive performance against strong open-weight baselines across diverse general and industrial coding benchmarks. Furthermore, the learned skills exhibit strong transferability and robust cross-domain generalization, highlighting the effectiveness of explicit behavioral abstraction for scalable agentic code generation.
\end{abstract}

\begin{figure*}[t]
  \centering
  \includegraphics[width=1\linewidth]{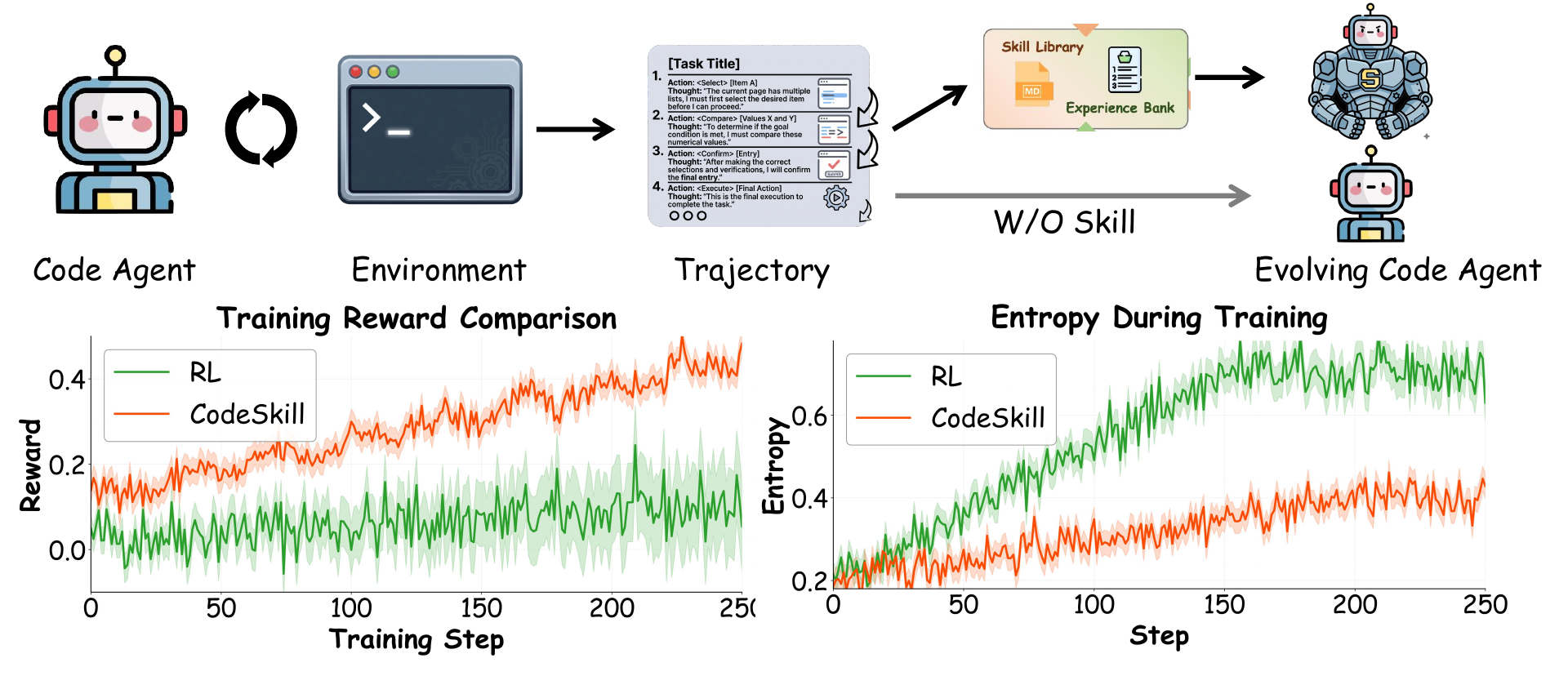}
  \caption{(a) CodeSkill pipeline overview. CodeSkill distills reusable skills from trajectories into a skill library, enabling efficient evolution of code agents, unlike vanilla training that directly learns from raw experience.
(b) Training comparison with Qwen2.5-Coder-32B-Instruct backbone. All runs use fixed seeds, repeated three times with std-based confidence intervals; rewards are normalized to identical scales. Vanilla RL refers to skill-free Proximal Policy Optimization (PPO). Curves are averaged over general and industrial code benchmarks. CodeSkill delivers faster reward convergence and lower exploration entropy, validating skill abstraction improves sample efficiency.}
  \label{fig:tea}
\end{figure*}

\section{Introduction}

Large language model (LLM)-based code agents are rapidly evolving from single-turn code generation toward agentic software engineering systems that iteratively navigate repositories, execute programs, edit files, and refine solutions through long-horizon interaction~\citep{jiang2026survey,joel2024survey,zhan2025kat}. Unlike static generation, these agents must perform sequential decision-making under complex environmental feedback, where intermediate actions can substantially influence future execution outcomes~\citep{wei2025supercoder,wu2026id}. Consequently, recent work has increasingly explored reinforcement learning (RL) with execution-based supervision, such as unit-test feedback, to optimize agent behaviors over multi-step trajectories~\citep{wen2025reinforcement,wu2026reaction}.

However, applying RL to code agents remains fundamentally challenging. Existing methods typically optimize directly over token-level trajectories, modeling long-horizon software engineering interactions as flat sequences of generation and tool-use actions. As illustrated in Figure~\ref{fig:tea}, this approach introduces two major limitations. First, sparse delayed rewards combined with large token-level action spaces often lead to unstable optimization and inefficient exploration. Second, flat trajectory modeling fails to capture reusable procedural structures across tasks. In practice, semantically similar behaviors, such as dependency tracing and iterative patch refinement, frequently recur across repositories despite differing surface, level trajectories. Consequently, these reusable patterns remain entangled within noisy, low-level action sequences, severely limiting long-horizon credit assignment and hindering effective experience reuse during RL training.

Recent agentic mid-training approaches have demonstrated that large-scale software engineering trajectories contain rich behavioral supervision signals~\citep{zeng2026davinci}. Yet, most existing methods still model these trajectories autoregressively at the token level, making it difficult to systematically abstract and reuse higher-level procedural behaviors. We hypothesize that this lack of explicit behavioral abstraction is an important factor underlying the instability of long-horizon RL for code agents. Without mechanisms that organize and reuse procedural experience, policies must repeatedly rediscover similar behaviors from low-level trajectories, substantially reducing exploration efficiency and generalization.

To address these limitations, we propose CodeSkill, a framework that adapts hierarchical latent skill modeling for the specific challenges of long-horizon code agents. Instead of optimizing purely over raw token trajectories, CodeSkill leverages a teacher model to extract recurring multi-step interaction patterns from both successful and failed rollouts. These trajectories are then distilled into a three-level hierarchy of textual abstractions: high-level global intent, mid-level execution skills, and low-level token control. To bridge the discrete semantic space of the LLM with the continuous exploration space of RL, CodeSkill employs variational inference to map these discrete textual skills into continuous latent variables. Each skill corresponds to a reusable procedural behavior, enabling the policy to optimize over semantically structured decision units in addition to low-level token generation. Furthermore, an adaptive boundary mechanism is introduced to dynamically gate skill transitions based on discrete execution feedback, ensuring temporal coherence. The learned skills are ultimately injected into frozen LLM policies as latent semantic prefixes.  

By shifting RL from token-level exploration to experience-level reasoning, CodeSkill explicitly improves optimization efficiency and long-horizon behavioral coherence. To rigorously evaluate our approach, we apply CodeSkill to a 32B-parameter backbone and test it across diverse general and industrial coding benchmarks. The primary contributions of this work are multifold. First, we introduce a hierarchical latent skill framework that successfully bridges discrete textual skill supervision with continuous RL optimization in coding environments. Second, we demonstrate that extracting counterfactual skills from both successful and failed trajectories provides robust supervisory signals that stabilize long-horizon optimization. Finally, empirical evaluations show that CodeSkill is highly competitive with strong, open-weight baselines across various benchmarks, demonstrating robust cross-domain generalization and strong transferability of the learned abstractions.

\section{Related Works}
\subsection{Reinforcement Learning and Mid-Training for LLMs}

Reinforcement learning (RL)~\citep{wen2025reinforcement,yu2025dapo,wang2026grpo} has become a key paradigm for improving reasoning in large language models by optimizing outcome-based rewards beyond supervised data. However, applying RL to multi-turn agentic tasks remains challenging due to the exponential growth of state space with interaction length~\citep{zhang2025agentrl}, leading to exploration collapse~\citep{wang2025arbitrary,wu2026capedit}, long-horizon credit assignment difficulty~\citep{feng2025group}, and fragile exploration under trajectory failure.
Mid-training (MT) bridges pre-training and RL by learning from offline interaction trajectories~\citep{yang2024swe,xu2024agenttrek}. Most existing methods rely on behavior cloning, which suffers from distribution shift~\citep{seo2024mitigating,guan2024explainable}, mode collapse~\citep{gx2025kl,wu2023re}, and performance bounded by demonstration quality~\citep{joel2024survey,jiang2026survey}, while also lacking temporal abstraction.
CodeSkill addresses these limitations by introducing a latent skill space learned from offline trajectories via variational inference. RL is then performed over skill-level actions instead of tokens, enabling temporally abstract exploration and more stable optimization. This design improves generalization and enhances sample efficiency in long-horizon code agent tasks.

\begin{figure*}[t]
  \centering
  \includegraphics[width=1\linewidth]{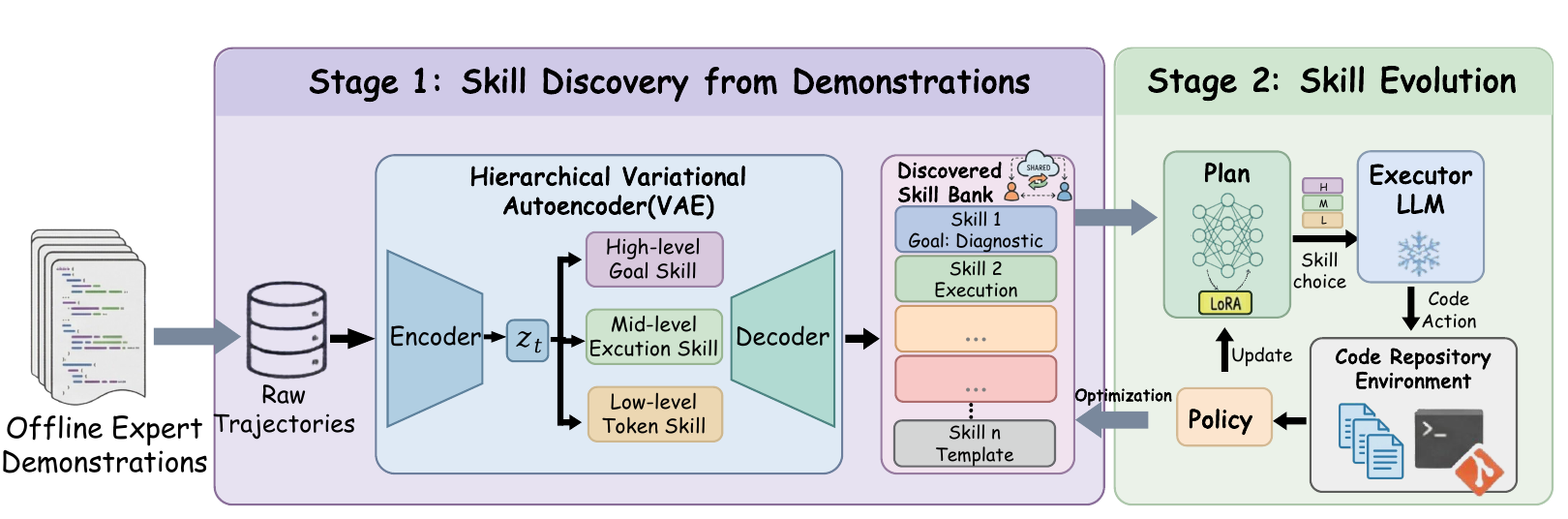}
  \caption{Overview of the CodeSkill framework. We present a two-stage framework for code agents: (1) Skill Discovery: A hierarchical VAE distills offline expert trajectories into a multi-level skill bank; (2) Skill Evolution: A planning-execution loop with RL refines skills based on environment feedback.}
  \label{fig:over}
\end{figure*}

\subsection{Agentic Skills}

A key direction for LLM agents is to introduce \emph{skills} as reusable abstractions for long-horizon decision making, elevating generation from token-level actions to higher-level behavioral primitives~\citep{wang2023voyager, xu2026agent, wu2026activemem}. Existing approaches construct skills via libraries or automatic discovery~\citep{alzubi2026evoskill, xia2026skillrl, zhang2026memskill}, but largely treat them as \emph{symbolic, discrete} units with heuristic boundaries, limiting their ability to capture the continuity and compositionality of behaviors and preventing joint optimization in skill space.
We adopt a \emph{latent-variable view}, modeling skills as continuous hierarchical representations, where skill discovery is cast as latent inference coupled with policy optimization. Accordingly, CodeSkill learns parameterized skills via variational inference with adaptive segmentation and integrates them into LLMs as soft conditioning, enabling unified optimization of skill abstraction and action generation.

\section{Method}

\subsection{Problem Formulation}

We formulate repository-level code generation as a sequential decision-making problem. Given a natural language instruction
$l \in \mathcal{L}$,
an agent interacts with an environment characterized by a state space
$\mathcal{S}$
and an action space
$\mathcal{A}$.
At timestep $t$, the state
$s_t \in \mathcal{S}$
contains repository context, execution feedback, file structures, and intermediate program states, while the action
$a_t \in \mathcal{A}$
corresponds to code edits, tool invocations, or execution operations.

We assume access to a dataset of repository interaction trajectories
$\mathcal{D}=\{(\tau_i,l_i)\}_{i=1}^{N}$,
where each trajectory
$\tau=\{(s_t,a_t)\}_{t=1}^{T}$
represents a multi-step software engineering process. 
Our objective is to learn a latent-conditioned policy
$\pi_\theta$, which acts simultaneously as the generative model and the reinforcement learning actor, that maximizes execution-based rewards while generalizing to unseen repository-level tasks. A comprehensive overview of the proposed framework designed to achieve this objective is depicted in Figure~\ref{fig:over}.

\textbf{Distinction from Traditional HRL.}
Traditional HRL~\citep{pateria2021hierarchical} learns latent skills from meaningless isotropic Gaussian priors $\mathcal{N}(0,I)$ and uses static probabilistic skill segmentation, which fails to align latent representations with coding semantics and ignores runtime execution feedback. When applied to long-horizon code agents, this misalignment disrupts complete programming subtasks and worsens sparse reward credit assignment.
By contrast, CodeSkill anchors continuous RL exploration to discrete textual coding semantics. Its temporal skill boundaries are dynamically gated by execution feedback like interpreter errors, unifying symbolic program states and continuous latent optimization to resolve code-specific long-horizon RL difficulties.

\subsection{Hierarchical Skill Discovery}

Raw repository-level trajectories are highly verbose, containing exploratory edits and redundant debugging attempts that obscure key decisions. Directly optimizing over such trajectories at the token level makes it difficult to identify reusable behavioral patterns. To address this, CodeSkill first transforms raw interaction trajectories into compact hierarchical skill representations.

We deploy a base code agent
$\pi_{\text{base}}$
to collect diverse trajectories:
$\mathcal{T}=\{\tau_i\}_{i=1}^{N}$.
Unlike prior approaches that retain only successful executions, we preserve both successful and failed trajectories:
\begin{equation}
\mathcal{T}^{+}=\{\tau_i:r(\tau_i)=1\},
\quad
\mathcal{T}^{-}=\{\tau_i:r(\tau_i)=0\},
\end{equation}
where
$r(\tau)$
denotes execution-based task success. Failed trajectories reveal debugging failure modes and execution boundary conditions critical for robust skill learning.

To avoid the confounding effect of knowledge distillation from a superior oracle, we employ the exact same backbone as $\pi_{\text{base}}$ for our teacher model $M_T$. $M_T$ is formulated via zero-shot structured prompting (see Appendix A for the full prompt templates) to distill hierarchical skills:
\begin{equation}
(s^H,s^M,s^L)=M_T(\tau,l).
\end{equation}

The high-level skill $s^H$ captures the global problem-solving strategy, mid-level skills $s^M=\{s_1^M,\dots,s_K^M\}$ represent temporally extended subtasks, and low-level skills $s^L$ capture localized execution behaviors. For $\mathcal{T}^{-}$, $M_T$ extracts compact counterfactual skills, summarizing the erroneous decision, the execution evidence and a corrected alternative.

\subsection{Skill Weaver}

The distilled hierarchical skills provide compact semantic descriptions of repository interactions, but they remain discrete textual representations that cannot be directly optimized by reinforcement learning. To bridge this gap, we introduce a Skill Weaver module, which parameterizes semantic Gaussian priors from the textual skills and maps them into continuous latent skill representations.

Specifically, at each timestep $t$, given a textual skill $s_t$, the frozen text embedding layer of the base LLM first encodes it into a semantic representation, which is then projected by a two-layer MLP to the parameters of a Gaussian distribution:
\begin{align}
p(z_t^H \mid s_t^H) &= \mathcal{N}(\mu_{\text{sem}}(s_t^H), \sigma_{\text{sem}}(s_t^H)), \\
p(z_t^L \mid s_t^L) &= \mathcal{N}(\mu_{\text{sem}}(s_t^L), \sigma_{\text{sem}}(s_t^L)),
\end{align}
where $\mu_{\text{sem}}(\cdot)$ and $\sigma_{\text{sem}}(\cdot)$ denote the mean and standard deviation of the semantic prior, respectively. They are parameterized by a two-layer MLP that projects the frozen LLM text embeddings from the hidden dimension $d_{\text{model}}$ to the latent space $d_z$. Consequently, semantically similar textual skills are encouraged to occupy nearby regions in the latent space, providing a semantically grounded prior for subsequent reinforcement learning.

For mid-level skills, we model temporal persistence through an \textit{execution-gated} boundary variable $b_t \in \{0, 1\}$. $p(b_t \mid s_t, z_t^H)$ is heavily biased by discrete state changes in $s_t$ (e.g., new compiler tracebacks). The mid-level prior is:
\begin{equation}
p(z_t^M \mid b_t, z_{t-1}^M, s_t^M) =
\begin{cases}
\mathcal{N}(\mu_{\text{sem}}(s_t^M), \sigma_{\text{sem}}(s_t^M)), & b_t=1, \\
\delta(z_t^M - z_{t-1}^M), & b_t=0.
\end{cases}
\end{equation}

\textbf{Variational Inference.} We employ an amortized variational posterior $q_\phi$. A bidirectional sequence encoder yields step-wise hidden states $h_t$ and a globally pooled state $h_{\text{global}}$. The posteriors are:
\begin{align}
q_\phi(z_t^H \mid h_{\text{global}}) &= \mathcal{N}(\mu^H(h_{\text{global}}), \sigma^H(h_{\text{global}})),\\
q_\phi(z_t^L \mid h_t) &= \mathcal{N}(\mu^L(h_t), \sigma^L(h_t)).
\end{align}

To allow joint backpropagation through the discrete boundary variable $b_t \sim q_\phi(b_t \mid h_t)$, we employ the Gumbel-Softmax relaxation during training. The mid-level posterior incorporates $b_t$:
\begin{equation}
q_\phi(z_t^M \mid h_t, b_t, z_{t-1}^M) =
\begin{cases}
\mathcal{N}(\mu_t^M(h_t),\sigma_t^M(h_t)), & b_t=1,\\
\delta(z_t^M-z_{t-1}^M), & b_t=0.
\end{cases}
\end{equation}

Finally, the Skill Weaver transforms the inferred latent skills into continuous soft prompts $P_t=\mathrm{Embed}(z_t^H,z_t^M,z_t^L)$, prepending them to the interaction history $\tilde H_t=\mathrm{Concat}(P_t,H_t)$. To align continuous soft prompts with the discrete token space without catastrophic forgetting, we freeze the base LLM and apply Low-Rank Adaptation (LoRA) to the attention layers, yielding the generation policy: $a_t\sim\pi_\theta(a_t\mid\tilde H_t)$.

\subsection{Variational Reinforcement Learning}

The variational objective forces the policy to stay close to the semantic priors:
\begin{equation}
\mathcal{L}_{\text{ELBO}} = \mathbb{E}_{q_\phi} \left[ \sum_{t=1}^{T} \log \pi_\theta(a_t\mid s_t,z_t^H,z_t^M,z_t^L) \right] - \mathcal{R}_{KL},
\end{equation}
\begin{align}
\mathcal{R}_{KL} &= \beta_H D_{KL}(q_\phi(z_t^H)\| p(z_t^H \mid s^H)) \notag\\
&\quad + \sum_{t=1}^{T} \Big[ \beta_B D_{KL}(q_\phi(b_t)\|p(b_t)) \notag\\
&\quad + b_t \cdot \beta_M D_{KL}(q_\phi(z_t^M)\| p(z_t^M \mid s_t^M)) \notag\\
&\quad + \beta_L D_{KL}(q_\phi(z_t^L)\| p(z_t^L \mid s_t^L)) \Big].
\end{align}

To align latent skills with repository-level success, we optimize the policy using Proximal Policy Optimization (PPO). Let $\hat{A}_t$ denote the generalized advantage estimation (GAE) based on the execution rewards. The RL objective is defined as:
\begin{equation}
\mathcal{L}_{RL} = \mathbb{E}_{\tau\sim\pi_\theta} \left[ \min(w_t \hat{A}_t, \text{clip}(w_t, 1-\epsilon, 1+\epsilon)\hat{A}_t) \right],
\end{equation}
where $w_t = \frac{\pi_\theta(a_t \mid \tilde H_t)}{\pi_{\text{old}}(a_t \mid \tilde H_t)}$. Crucially, for failed trajectories $\tau \in \mathcal{T}^{-}$, the counterfactual skills extracted by $M_T$ guide the policy toward negative advantages ($\hat{A}_t < 0$), explicitly penalizing degenerate latent behaviors. The final loss is:
\begin{equation}
\mathcal{L} = \mathcal{L}_{\text{ELBO}} + \lambda \mathcal{L}_{RL}.
\end{equation}

\section{Experiments}
\begin{table*}[htbp]
\centering
\caption{Performance comparison on general code generation tasks.}
\resizebox{0.98\textwidth}{!}{%
\begin{tabular}{l c | c c c c | c c | c}
\toprule
\multirow{2}{*}{Model} & \multirow{2}{*}{Size} & \multicolumn{4}{c|}{EvalPlus} & \multicolumn{2}{c|}{BigCodeBench} & \multirow{2}{*}{FullStackBench} \\
& & HumanEval & HumanEval+ & MBPP & MBPP+ & Full & Hard & \\
\midrule
\multicolumn{9}{c}{\textbf{6B+ Models}} \\
\midrule
DeepSeek-Coder-V2-Lite-Instruct & 2.4/16B & 81.1 & 75.6 & 85.2 & 70.6 & 37.8 & 18.9 & 49.4 \\
Qwen2.5-Coder-7B-Instruct & 7B & 87.2 & 81.7 & 84.7 & 72.2 & 37.8 & 13.5 & 42.2 \\
Seed-Coder-8B-Instruct & 8B & 81.1 & 75.6 & 86.2 & 73.3 & 44.6 & 23.6 & 55.8 \\
Qwen2.5-Coder-14B-Instruct & 14B & 62.8 & 59.8 & 88.6 & 77.2 & 47.0 & 6.1 & 53.1 \\
\midrule
\multicolumn{9}{c}{\textbf{30B+ Models}} \\
\midrule
Qwen3-Coder-30B-A3B-Instruct & 3.3/30.5B & 93.9 & 87.2 & 90.7 & 77.2 & 46.9 & 27.7 & 60.9 \\
Deepseek-V3.2 & 37/671B & 93.9 & 88.4 & 93.4 & 77.2 & 48.1 & 27.0 & 64.9 \\
Qwen2.5-Coder-32B-Instruct & 32B & 93.3 & 86.6 & 90.2 & 77.8 & 48.0 & 24.3 & 57.4 \\
Qwen3-235B-A22B-Instruct-2507 & 22/235B & 96.3 & 91.5 & 92.3 & 77.8 & 47.4 & 25.7 & 62.7 \\
Qwen3-235B-A22B-Thinking-2507 & 22/235B & 98.8 & 93.3 & 95.5 & 81.5 & 44.1 & 23.0 & - \\
Qwen3-Coder-480B-A35B-Instruct & 35/480B & 97.6 & 92.7 & 94.2 & 80.2 & 49.4 & 27.7 & 66.4 \\
Kimi-Dev-72B & 72B & 93.3 & 86.0 & 79.6 & 68.8 & 45.4 & 31.8 & 38.6 \\
Kimi-K2-Instruct-0905 & 32B/1T & 94.5 & 89.6 & 91.8 & 74.1 & 49.8 & 30.4 & 63.5 \\
Kimi-K2-Thinking & 32B/1T & 98.2 & 92.7 & 97.4 & 82.3 & 46.8 & 28.4 & - \\
KAT-Dev & 32B & 90.9 & 86.6 & 89.4 & 76.2 & 46.2 & 25.7 & 58.8 \\
KAT-Dev-72B-Exp & 72B & 88.4 & 81.7 & 85.2 & 69.3 & 48.3 & 26.4 & 52.9 \\
GLM-4.7 & 32/355B & 87.2 & 79.9 & 90.5 & 75.7 & 45.7 & 26.4 & 70.2 \\
InCoder-32B & 32B & 94.5 & 89.6 & 91.8 & 78.3 & 49.8 & 31.1 & 57.1 \\
InCoder-32B-Thinking & 32B & 95.1 & 89.6 & 92.1 & 78.3 & 47.4 & 29.1 & 60.8 \\
CodeSkill & 32B & \textbf{99.2} & \textbf{94.8} & \textbf{98.3} & \textbf{83.7} & \textbf{50.3} & \textbf{32.6} & \textbf{71.5} \\
\bottomrule
\end{tabular}%
}
\label{tab:code_generation}
\end{table*}
\begin{table*}[htbp]
\centering
\caption{Performance comparison on chip design benchmarks. CodeSkill results are highlighted in blue.}
\resizebox{\linewidth}{!}{%
\begin{tabular}{lr|c|c|cc|cccc|cc}
\toprule
\multirow{2}{*}{Model} & \multirow{2}{*}{Size} & \multirow{2}{*}{\makecell{VeriScope \\ Score}} & \multirow{2}{*}{\makecell{VeriRepair \\ Fix (\%)}} & \multicolumn{6}{c|}{RealBench} & \multicolumn{2}{c}{ArchXBench} \\
\cline{5-12}
& & & & \multicolumn{2}{c|}{System} & \multicolumn{4}{c|}{Module} & $n$ & $t$ \\
& & & & Syn@1 & Syn@5 & Syn@1 & Syn@5 & Func@1 & Func@5 & & \\
\midrule
\multicolumn{12}{c}{6B+ Models} \\
\midrule
Qwen3.5-9B & 9B & 32.0 & - & - & - & 6.3 & 15.8 & 4.3 & 8.5 & 1.9 & 44.3 \\
GPT-OSS-20B & 3.6/21B & 73.9 & 86.7 & 3.8 & 17.4 & 22.9 & 47.9 & 9.8 & 21.0 & 3.1 & 53.5 \\
Qwen3.5-27B & 27B & 55.7 & 60.0 & 6.2 & 20.1 & 17.1 & 33.8 & \textbf{10.6} & 17.8 & 2.6 & 48.3 \\
\midrule
\multicolumn{12}{c}{30B+ Models} \\
\midrule
Qwen3-Coder-30B-A3B-Instruct & 3.3/30.5B & 66.2 & 76.7 & - & - & 23.0 & 35.2 & 5.2 & 8.2 & 2.4 & 37.3 \\
Seed-OSS-36B-Instruct & 36B & 67.2 & 66.7 & 5.0 & 21.3 & 14.3 & 23.0 & 11.5 & 20.3 & 2.5 & 43.7 \\
GPT-OSS-120B & 5.1/117B & 82.2 & 76.7 & 5.0 & 21.3 & 37.8 & 64.3 & 17.5 & 30.8 & 3.4 & 54.8 \\
MiniMax-M2.5 & 10/230B & 75.1 & 66.7 & 23.8 & 48.5 & 17.2 & 38.4 & 6.9 & 15.9 & 2.9 & 46.0 \\
GLM-4.7 & 32/355B & 81.2 & 63.3 & 12.5 & 24.6 & 25.4 & 46.9 & 11.6 & 21.2 & 3.2 & 51.4 \\
GLM-5 & 40/744B & 83.2 & 90.0 & 2.5 & 11.2 & 22.2 & 43.4 & 12.2 & 22.6 & 3.1 & 53.2 \\
Kimi-K2.5 & 32B/1T & 73.1 & 83.3 & 5.0 & 17.9 & 43.7 & 52.2 & 23.1 & 25.7 & 3.8 & 49.7 \\
Kimi-K2-Instruct & 32B/1T & 82.4 & 76.7 & 6.2 & 26.2 & 50.1 & 70.1 & 22.2 & 28.3 & 2.9 & 44.9 \\
Kimi-K2-Thinking & 32B/1T & 73.1 & 80.0 & - & - & 27.8 & 59.4 & 14.1 & 28.9 & 1.5 & 30.1 \\
DeepSeek-V3.2 & 37/671B & 76.1 & 77.0 & 18.8 & 55.1 & 39.3 & 52.7 & 17.2 & 21.4 & 3.6 & 53.9 \\
Qwen3.5-397B-A17B & 17/397B & 62.5 & 86.7 & 11.2 & 38.1 & 35.2 & 59.5 & 16.4 & 28.3 & 3.1 & 53.5 \\
Qwen3-Coder-480B-A35B-Instruct & 35/480B & 80.8 & 76.7 & - & - & 28.9 & 39.5 & 14.8 & 20.6 & 3.0 & 43.9 \\
InCoder-32B & 32B & 80.7 & 80.0 & 10.0 & 23.7 & 74.8 & 83.3 & 62.7 & 70.5 & 3.4 & 51.0 \\
InCoder-32B-Thinking & 32B & 75.4 & 83.3 & 12.3 & 24.5 & 75.6 & 82.4 & 63.1 & 69.8 & 3.1 & 46.7 \\
CodeSkill & 32B & \textbf{89.6} & \textbf{90.4} & \textbf{43.8} & \textbf{56.4} & \textbf{78.8} & \textbf{85.7} & \textbf{66.2} & \textbf{73.6} & \textbf{4.8} & \textbf{59.5}\\
\midrule
\multicolumn{12}{c}{Closed-APIs Models} \\
\midrule
Claude-Sonnet-4.6 & \faLock &87.7 & 83.3 & 41.2 & 50.0 & 69.2 & 77.7 & 33.5 & 37.2 & 4.4 & 58.2 \\
\bottomrule
\end{tabular}%
}
\label{tab:chip_design}
\end{table*}
\subsection{Baselines and Benchmarks}

We evaluate CodeSkill against a diverse set of state-of-the-art coding models, including both open-weight models and proprietary APIs~\citep{qwen3dot52026,bytedance2025seedoss,qwen3coder2025,zhipu2025glm,anthropic2026claude46,cao2026qwen3,liu2025deepseek,zhu2024deepseek,team2025kimi,hui2024qwen2,yang2026incoder,yang2025qwen3,seed2025seed,zhan2025kat}. Detailed baseline configurations are provided in the Appendix.
The evaluation covers both general-purpose and industrial repository-level coding benchmarks~\citep{liu2023your,yu2018spider,li2023can,chen2021evaluating,du2024mercury,jain2025livecodebench,gu2024cruxeval,cheng2024fullstack,zhuo2025bigcodebench,austin2021program}, with detailed benchmark descriptions provided in the Appendix. Among them, Mercury~\citep{du2024mercury} is adopted to specifically evaluate code efficiency, as it jointly measures functional correctness and runtime performance.
Unless otherwise specified, CodeSkill is built upon the Qwen2.5-Coder-32B-Instruct backbone.

\begin{table*}[htbp]
\centering
\caption{Performance on GPU optimization, code optimization, and 3D modeling benchmarks. InCoder-32B-Thinking results are highlighted in gray.}
\resizebox{\linewidth}{!}{%
\begin{tabular}{lr|cc|c|cc|cccc|ccc}
\toprule
\multirow{2}{*}{Model} & \multirow{2}{*}{Size} & \multicolumn{2}{c|}{CAD-Coder} & \multirow{2}{*}{\makecell{EmbedCGen \\ Main (\%)}} & \multicolumn{2}{c|}{SuperCoder} & \multicolumn{4}{c|}{TritonBench} & \multicolumn{3}{c}{KernelBench} \\
\cline{3-5} \cline{6-7} \cline{8-11} \cline{12-14}
& & Comp. & IoU & & Acc. (\%) & Spd. & G-call (\%) & G-exe (\%) & T-call (\%) & T-exe (\%) & L1 & L2 & L3 \\
\midrule
\multicolumn{14}{c}{6B+ Models} \\
\midrule
Qwen3.5-9B & 9B & 2.0 & - & 10.0 & 36.0 &  \textbf{1.0×} & 2.7 & 100.0 & 3.6 & 100.0 & - & - & - \\
GPT-OSS-20B & 3.6/21B & 2.0 & 2.0 & 30.6 & 16.0 & 1.0× & 2.2 & 100.0 & 1.2 & 100.0 & 5.1 & 10.0 & 2.0 \\
Qwen3.5-27B & 27B & 4.0 &  0.3 & 9.6 & 6.0 & 1.9× & 5.4 & 100.0 & 25.9 & 97.7 & 5.1 & 6.0 & 4.0 \\
\midrule
\multicolumn{14}{c}{30B+ Models} \\
\midrule
Qwen3-Coder-30B-A3B-Instruct & 3.3/30.5B & - & - & 15.4 & 50.0 & 1.0× & 8.7 & 100.0 & 24.1 & 67.5 & - & - & - \\
Seed-OSS-36B-Instruct & 36B & 2.0 & 2.0 & 10.2 & 8.0 & 1.1× & 1.6 & 100.0 & 1.8 & 100.0 & 1.0 & 2.0 & 2.0 \\
GPT-OSS-120B & 5.1/117B & 4.0 & 1.9 & 17.8 & 8.0 & 1.2× & 3.8 & 85.7 & 12.7 & 95.2 & 6.1 & 15.0 & 2.0 \\
MiniMax-M2.5 & 10/230B & 4.0 & 0.4 & 22.2 & 20.0 & 3.5× & 5.4 & 100.0 & 15.1 & 100.0 & 7.1 & 14.0 & 8.0 \\
GLM-4.7 & 32/355B & 12.0 & 6.0 & 89.6 & 20.0 & 8.6× & 3.3 & 100.0 & 6.0 & 100.0 & 8.1 & 19.0 & - \\
GLM-5 & 40/744B & 38.0 & 18.8 &  90.2 & 54.0 & 1.87× & 1.6 & 100.0 & 1.2 & 100.0 & 16.2 & 23.0 & 4.0 \\
Kimi-K2.5 & 32B/1T & 40.0 & 12.1 & 81.0 & 46.0 & 1.9× & 12.5 & 100.0 & 7.8 & 100.0 & 13.1 & 23.0 & 6.0 \\
Kimi-K2-Instruct & 32B/1T & 2.0 & 1.1 & 69.6 & 12.0 & 1.1× & 15.8 & 96.5 & 13.9 & 91.3 & 6.1 & - & - \\
Kimi-K2-Thinking & 32B/1T & 48.0 & 20.0 & - & 24.0 & 1.2× & 17.4 & 100.0 & 19.9 & 84.8 & 9.1 & 16.0 & 4.0 \\
DeepSeek-V3.2 & 37/671B & 14.0 & 4.6 & 84.4 & 30.0 & 1.8× & 19.6 & 100.0 & 18.1 & 13.3 & 3.0 & - & - \\
Qwen3.5-397B-A17B & 17/397B & 36.0 & 14.2 & 17.8 & 34.0 & 1.2× & 7.6 & 100.0 & 16.3 & 92.6 & 4.0 & 10.0 & - \\
Qwen3-Coder-480B-A35B-Instruct & 35/480B & 10.0 & 4.7 & 9.0 & 64.0 & 2.0× & 20.1 & 100.0 & 31.9 & 56.6 & 3.0 & 6.0 & - \\
InCoder-32B & 32B & 82.0 & 53.5 & 35.2 & 91.0 & 1.3× & 18.5 & 100.0 & 19.3 & 93.8 & 22.2 & 36.0 & 14.0 \\
InCoder-32B-Thinking & 32B & 84.0 & 48.6 & 47.9 & 93.0 & 3.93× & 15.2 & 100.0 & 22.9 & 89.5 & 20.2 & 38.0 & 12.0 \\
CodeSkill & 32B &\textbf{87.0} &\textbf{56.3} &\textbf{93.4} & \textbf{95.0} & 1.2× &\textbf{28.4} & 100.0 & \textbf{40.7} & 100.0 &  \textbf{25.2} &  \textbf{41.0}&  \textbf{15.0} \\
\midrule
\multicolumn{14}{c}{Closed-APIs Models}\\
\midrule
Claude-Sonnet-4.6 & \faLock & 77.0 & 32.4 &  79.0 & 88.0 & 4.6× &  26.4 & 98.1 &  36.8 & 1.4 & 11.1 & 28.0 & 2.0 \\
\bottomrule
\end{tabular}%
}
\label{tab:gpu_opt}
\end{table*}

\subsection{Main Results}
\paragraph{Results on General Code Benchmarks.}

As shown in Table~\ref{tab:code_generation}, CodeSkill consistently demonstrates strong and well-balanced performance across a broad range of coding tasks, including code generation, reasoning, and interactive tool use. Despite operating at the 32B scale, the model remains highly competitive with substantially larger systems and achieves leading performance among open-weight models on several challenging benchmarks.
The gains are particularly evident on tasks requiring long-horizon reasoning, structured problem decomposition, and execution-aware decision making, suggesting that latent skill abstraction improves the model’s ability to maintain coherent multi-step behaviors beyond conventional token-level generation. In addition, CodeSkill exhibits strong robustness across both general-purpose coding benchmarks and specialized agentic environments, indicating that the learned latent skills transfer effectively across heterogeneous software engineering tasks.
Overall, the results suggest that CodeSkill provides a scalable and effective framework for enhancing code agents, enabling strong generalization across realistic coding scenarios while maintaining competitive performance efficiency at moderate model scale.

\begin{table}[t]
\centering
\caption{Resolve rate (pass@1) on benchmarks. Results are mean $\pm$ std over three runs. For each model, the best result is highlighted in gray.}
\resizebox{0.48\textwidth}{!}{%
\begin{tabular}{l|ccccc}
\toprule
Model & Origin & SFT & \textsc{PPO} & \textsc{GRPO} & \text{CodeSkill} \\
\midrule
\multicolumn{4}{c}{\textsc{SWEBench-Verified}} \\
\midrule
Qwen3-4B & 1.7 ($\pm$ 0.1) & 2.4 ($\pm$ 0.3) & 4.9 ($\pm$ 0.6) & 7.3 ($\pm$ 0.8)  & 14.3 ($\pm$ 0.2) \\
Gemma-3-27b & 7.1 ($\pm$ 0.5) & 7.0 ($\pm$ 0.4) & 9.3 ($\pm$ 0.8) & 11.6 ($\pm$ 0.5) &  19.7 ($\pm$ 0.3) \\
Qwen3-Coder-30B & 37.7 ($\pm$ 0.2) & 43.8 ($\pm$ 0.8) & 48.1 ($\pm$ 0.9) & 49.5 ($\pm$ 0.6)  &  58.7 ($\pm$ 0.5) \\
GLM-4.5-Air & 51.4 ($\pm$ 0.2) & 51.5 ($\pm$ 0.8) & 52.9 ($\pm$ 0.6) & 54.2 ($\pm$ 0.7)  &   62.8 ($\pm$ 0.3) \\
\midrule
\multicolumn{4}{c}{\textsc{SWEBench-Lite}} \\
\midrule
Qwen3-4B & 1.2 ($\pm$ 0.3) & 1.3 ($\pm$ 0.5) & 2.4 ($\pm$ 0.8) & 6.3 ($\pm$ 0.5) &  12.9 ($\pm$ 0.2) \\
Gemma-3-27b & 6.0 ($\pm$ 0.8) & 5.9 ($\pm$ 0.4) & 7.0 ($\pm$ 0.4) & 8.8 ($\pm$ 0.9) &  15.3 ($\pm$ 0.3) \\
Qwen3-Coder-30B & 28 ($\pm$ 0.3) & 33.9 ($\pm$ 0.3) & 41.4 ($\pm$ 0.7) & 42.9 ($\pm$ 0.9)  & 51.7 ($\pm$ 0.4) \\
GLM-4.5-Air & 43.5 ($\pm$ 0.6) & 43.9 ($\pm$ 0.4) & 44.5 ($\pm$ 0.8) & 47.2 ($\pm$ 0.8) &  53.4 ($\pm$ 0.3) \\
\bottomrule
\end{tabular}
}
\label{tab:main_result}
\end{table}

\begin{table*}[htbp]
\centering
\caption{Ablation study on general code generation tasks. To ensure fair comparisons, the textual skill baselines are strictly matched with the latent variants in terms of tunable parameter count and prompt context length.}
\resizebox{0.9\textwidth}{!}{%
\begin{tabular}{l | c c c c | c c | c}
\toprule
\multirow{2}{*}{Model}  & \multicolumn{4}{c|}{EvalPlus} & \multicolumn{2}{c|}{BigCodeBench} & \multirow{2}{*}{FullStackBench} \\
& HumanEval & HumanEval+ & MBPP & MBPP+ & Full & Hard & \\
\midrule
CodeSkill & \textbf{99.2} & \textbf{94.8} & \textbf{98.3} & \textbf{83.7} & \textbf{50.3} & \textbf{32.6} & \textbf{71.5} \\
\midrule
\multicolumn{8}{l}{\textit{Hierarchy Ablations}} \\
w/o hierarchy  & 95.8 & 88.7 & 92.1 & 78.4 & 46.9 & 28.5 & 61.8 \\
w/o $z_H$  & 96.4 & 89.6 & 92.8 & 79.1 & 47.8 & 29.4 & 62.6 \\
w/o $z_M$  & 95.9 & 88.9 & 92.0 & 78.2 & 46.5 & 27.9 & 61.3 \\
w/o $z_L$  & 96.8 & 90.5 & 93.1 & 79.8 & 48.6 & 30.7 & 63.4 \\
\midrule
\multicolumn{8}{l}{\textit{VAE \& Modeling Ablations}} \\
deterministic posterior (mean only)  & 96.2 & 89.4 & 92.5 & 78.7 & 47.3 & 28.8 & 62.1 \\
no boundary (fixed segmentation)  & 96.5 & 90.1 & 92.9 & 79.5 & 48.1 & 29.9 & 62.8 \\
\midrule
\multicolumn{8}{l}{\textit{Learning Mechanism \& Decomposition Ablations}} \\
token-level RL (w/o skills)  & 96.1 & 89.2 & 92.3 & 78.6 & 47.0 & 28.4 & 61.7 \\
CodeSkill (SFT only, w/o RL)  & 95.4 & 87.9 & 91.5 & 77.2 & 45.3 & 26.4 & 60.1 \\
textual skills + RL (w/o VAE)  & 95.6 & 88.4 & 91.8 & 77.9 & 46.1 & 27.5 & 60.9 \\
textual skills (SFT only, w/o RL \& VAE) & 94.5 & 86.2 & 90.4 & 76.3 & 44.0 & 25.4 & 58.3 \\
\midrule
\multicolumn{8}{l}{\textit{Data Ablations}} \\
success-only trajectories  & 96.7 & 90.3 & 93.0 & 79.7 & 48.4 & 30.1 & 63.0 \\
failure-only trajectories  & 93.1 & 84.8 & 89.7 & 74.2 & 42.6 & 23.1 & 57.5 \\
\bottomrule
\end{tabular}%
}
\label{tab:ab_code_generation}
\end{table*}

\paragraph{Results on Industrial Code Benchmarks.}
As shown in Table~\ref{tab:chip_design} and Table~\ref{tab:gpu_opt}, CodeSkill consistently demonstrates strong generalization across diverse industrial coding domains, including chip design, GPU optimization, 3D modeling, and low-level kernel synthesis. Compared with existing open-weight coding models, CodeSkill achieves more stable and competitive performance across both functional correctness and system-level engineering tasks, highlighting the effectiveness of latent skill abstraction for long-horizon technical reasoning.
On chip design benchmarks, CodeSkill attains the strongest overall open-weight performance, particularly on challenging repository- and module-level synthesis tasks such as RealBench, while remaining highly competitive across VeriScope, VeriRepair, and ArchXBench. The gains are especially pronounced on complex module synthesis and functional verification metrics, suggesting that hierarchical latent skills improve structured reasoning over long dependency chains and multi-stage hardware generation workflows.
CodeSkill also exhibits clear advantages on GPU optimization and 3D engineering tasks. The model achieves leading open-weight performance on CAD-Coder and KernelBench, indicating strong capability in geometric reasoning, low-level kernel generation, and execution-aware optimization. In particular, the consistent improvements across increasing KernelBench difficulty levels suggest that the learned latent skills help stabilize long-horizon optimization and code refinement processes beyond standard next-token generation.
Overall, the results demonstrate that CodeSkill generalizes effectively beyond conventional functional code generation benchmarks, providing robust performance across a wide range of industrial engineering scenarios that require structured planning, iterative refinement, and execution-aware reasoning.

\subsection{Ablation Study}

We conduct comprehensive ablation experiments across two evaluation suites to systematically validate the efficacy of CodeSkill's core components. For Table~\ref{tab:main_result}, which targets complex repository-level SWE agent tasks, we report the pass@1 resolution rates alongside standard deviations over three independent trials. Across all evaluated model backbones on both SWE-bench Verified and SWE-bench Lite, CodeSkill consistently outperforms the raw base models, vanilla SFT, and standard token-level RL algorithms such as PPO and GRPO. This performance edge is fundamentally derived from our hierarchical latent skill design. Instead of relying on flat, token-wise trajectory modeling that exacerbates sparse reward credit assignment errors over long horizons, CodeSkill abstracts reusable, multi-step coding subtasks into structured, continuous latent variables. By optimizing over these semantically rich latent spaces, CodeSkill mitigates the inefficient exploration inherent to token-level RL, accelerates valid behavioral discovery, and effectively leverages counterfactual signals from failed debugging paths to prevent repetitive invalid operations. Consequently, it successfully overcomes the behavioral incoherence that plagues traditional token-wise baselines over extended interaction lengths.

Table~\ref{tab:ab_code_generation} further breaks down the contribution of each core module within CodeSkill on general code benchmarks. The degradation in performance upon the removal of any single component underscores that hierarchical latent skill modeling and reinforcement learning must operate collaboratively. Ablating the multi-level hierarchy entirely precipitates the most severe performance decline, particularly on complex sequential benchmarks like BigCodeBench and FullStackBench. Within this hierarchy, the mid-level latent variable $z_M$ proves to be the most critical element; its removal disproportionately destabilizes temporally extended subtasks, confirming its role in maintaining mid-to-long range execution coherence. Additionally, the ablation results demonstrate that both stochastic VAE sampling and the execution-triggered adaptive boundary mechanism are strictly necessary. Reverting to a deterministic posterior, which utilizes only the mean, or employing fixed trajectory segmentation noticeably impairs the model's ability to dynamically gate skill transitions based on runtime execution feedback.
To explicitly disentangle the benefits of offline skill decomposition from RL optimization, we construct baselines with strictly matched parameter budgets and prompt lengths. Applying SFT exclusively on discrete textual skills yields only marginal improvements, and while augmenting these textual skills with RL delivers further gains, it still significantly underperforms the full CodeSkill framework. This comparison validates that bridging discrete semantics into a continuous VAE latent space is indispensable for efficient and stable RL exploration. Furthermore, SFT-only training within the latent space without RL refinement underperforms, confirming that reward-based RL is mandatory to align the latent skills with repository-level success. Finally, our data ablations reveal that training on a mixture of successful and failed trajectories yields peak performance. Although training exclusively on failed trajectories induces severe behavioral regression, strategically balancing them with successful samples provides the policy with crucial counterfactual skills. This integration explicitly penalizes degenerate latent behaviors and significantly enhances the robustness of the learned skills against complex execution boundary conditions.

\section{Conclusions}

In this work, we presented CodeSkill, a novel framework for repository-level code generation that shifts reinforcement learning from low-level token optimization to high-level skill reasoning. By abstracting historical interaction trajectories into hierarchical semantic skills and grounding policy optimization in a structured latent space, CodeSkill enables more effective long-horizon exploration while leveraging the reasoning capabilities of the underlying LLM. Extensive experiments on both public and industrial benchmarks demonstrate consistent improvements in repository-level code generation performance and long-horizon decision making. Moreover, the strong transferability and cross-domain generalization of the learned hierarchical skills suggest that semantic skill abstraction provides a promising foundation for scalable and robust code agents.

\bibliography{aaai2027}

\section{Benchmark Details}
\subsection{Baselines}
\label{baseline}
To ensure the integrity of our evaluation and strictly rule out data contamination, we conducted rigorous checks, confirming zero data overlap among the training corpora (SWE-Smith and daVinci-Dev), the constructed skill library, and the final evaluation benchmarks. Our proposed CodeSkill model is trained based on the frozen Qwen2.5-Coder-32B-Instruct backbone with LoRA fine-tuning and latent skill reinforcement learning.

Because general-code and industrial-code benchmarks target distinct skill sets, we utilize tailored baseline configurations for each setting, ensuring that our reference models are highly competitive and relevant to each benchmark category.

For the general-code evaluations, CodeSkill is compared alongside a robust selection of models. These include DeepSeek-Coder-V2-Lite-Instruct~\citep{zhu2024deepseek} and DeepSeek-V3.2~\citep{liu2025deepseek}; the Qwen2.5-Coder series (7B, 14B, and 32B variants)~\citep{hui2024qwen2}; Qwen3-235B-A22B-Instruct-2507 and Qwen3-235B-A22B-Thinking2507~\citep{yang2025qwen3}; Qwen3-Coder-30B-A3B-Instruct and Qwen3-Coder-480B-A35B-Instruct~\citep{qwen3coder2025}; SeedCoder-8B-Instruct~\citep{seed2025seed}; Kimi-Dev-72B~\citep{yang2025kimi}; Kimi-K2-Instruct-0905 and Kimi-K2-Thinking~\citep{team2025kimi}; KAT-Dev and KAT-Dev-72B-Exp~\citep{zhan2025kat}; GLM-4.7~\citep{zhipu2025glm}; and InCoder-32B~\citep{yang2026incoder}. By spanning a wide spectrum of parameter scales and incorporating both dense and mixture-of-experts (MoE) architectures, this ensemble provides a rigorous comparative standard for measuring general coding capabilities.

For industrial-code evaluations, we deploy a specialized set of baselines curated specifically for their relevance to hardware-aware code generation, program optimization, and other domain-specific engineering tasks. Our comparison encompasses a robust lineup of models, including the Qwen3.5 series (9B, 27B, and 397B-A17B)~\citep{qwen3dot52026}; the Qwen3-Coder family (30B-A3B and 480B-A35B)~\citep{qwen3coder2025,cao2026qwen3}; DeepSeek-V3.2~\citep{liu2025deepseek}; GLM-4.7~\citep{zhipu2025glm} and GLM5~\citep{zeng2026glm}; Kimi-K2.5~\citep{team2026kimi}, alongside Kimi-K2-Instruct and Kimi-K2-Thinking~\citep{team2025kimi}; MiniMax-M2.5~\citep{minimaxm252026}; Seed-OSS-36B-Instruct~\citep{bytedance2025seedoss}; and the GPT-OSS models (20B and 120B)~\citep{agarwal2025gpt}. Furthermore, we incorporate ClaudeSonnet-4.6~\citep{anthropic2026claude46} to serve as a strong proprietary reference. This carefully selected ensemble ensures a comprehensive evaluation of CodeSkill against the leading systems in specialized industrial applications.

\subsection{Benchmarks}
\label{Benchmarks}
\paragraph{General Code Benchmarks.} Following the established evaluation protocol of CodeSkill, we assess CodeSkill across a comprehensive suite of general code benchmarks. This suite is designed to test a wide array of capabilities, including code generation, reasoning, efficiency, text2SQL translation, agentic coding, and tool utilization. The evaluated benchmarks comprise EvalPlus~\citep{liu2023your} (incorporating HumanEval~\citep{chen2021evaluating} and MBPP~\citep{austin2021program}), BigCodeBench~\citep{zhuo2025bigcodebench}, FullStackBench~\citep{cheng2024fullstack}, CRUXEval~\citep{gu2024cruxeval}, LiveCodeBench~\citep{jain2025livecodebench}, Mercury~\citep{du2024mercury}, Spider~\citep{yu2018spider}, BIRD~\citep{li2023can}, Terminal-Bench~\citep{teamterminal}, SWE-bench Verified~\citep{swebenchverified}, Mind2Web~\citep{deng2023mind2web}, BFCL V3~\citep{patil2025berkeley}, and $\tau^2$-bench~\citep{barres2025tau}. Detailed descriptions of these benchmarks and the corresponding evaluation methodologies are provided in Appendix.

\paragraph{Industrial Code Benchmarks.} Our primary focus lies in the evaluation of industrial coding capabilities. To this end, we employ the rigorous benchmark suite introduced with CodeSkill, which targets hardware-oriented and engineering-intensive programming tasks. Specifically, we report performance on VeriScope, RealBench~\citep{jin2025realbench}, ArchXBench~\citep{purini2025archxbench}, and VeriRepair for chip design; KernelBench~\citep{ouyang2025kernelbench} and TritonBench~\citep{li2025tritonbench} for GPU kernel optimization; EmbedCGen and SuperCoder~\citep{wei2025supercoder} for low-level code optimization; and CAD-Coder~\citep{guan2026cad} for 3D modeling. These benchmarks present a substantially higher degree of difficulty compared to standard software tasks. They necessitate that models adhere to strict compilation and simulation constraints, reason deeply about hardware architecture and performance trade-offs, and generate outputs validated by specialized, domain-specific pipelines. Consequently, these benchmarks serve as our primary testbed to ascertain CodeSkill's efficacy in complex, real-world industrial coding scenarios.

\begin{table*}[htbp]
\centering
\caption{Performance comparison on agentic coding tasks (Terminal-Bench v1.0, Terminal-Bench v2.0, SWE-Verified) and general tool-use tasks (Mind2Web, BFCL V3, $\tau^2$-bench).}
\resizebox{\textwidth}{!}{%
\begin{tabular}{l c | c c c | c c c c c}
\toprule
\multirow{3}{*}{Model} & \multirow{3}{*}{Size} & \multicolumn{3}{c|}{Agentic Coding} & \multicolumn{5}{c}{General Tool Use} \\
\cmidrule{3-10}
& & \multicolumn{2}{c}{Terminal-Bench} & \multirow{2}{*}{SWE-bench Verified} &  \multirow{2}{*}{Mind2Web} &  \multirow{2}{*}{BFCL V3} & \multicolumn{3}{c}{$\tau^2$-bench} \\
& & v1.0 & v2.0 & & & & Airline & Retail & Telecom \\
\midrule
\multicolumn{10}{c}{\textbf{6B+ Models}} \\
\midrule
DeepSeek-Coder-V2-Lite-Instruct & 2.4/16B & 5.0 & - & - & 26.7 & - & 3.5 & 12.0 & - \\
Qwen2.5-Coder-7B-Instruct & 7B & 6.3 & - & - & 38.4 & 54.2 & - & - & - \\
Seed-Coder-8B-Instruct & 8B & 7.5 & 2.5 & - & 38.2 & - & 4.3 & 32.0 & - \\
Qwen2.5-Coder-14B-Instruct & 14B & 8.8 & - & - & 42.7 & 59.9 & - & - & - \\
\midrule
\multicolumn{10}{c}{\textbf{30B+ Models}} \\
\midrule
Qwen3-Coder-30B-A3B-Instruct & 3.3/30.5B & 23.8 & 23.8 & 51.9 & 36.1 & 63.4 & 42.0 & 25.4 & 25.4 \\
DeepSeek-v3.2 & 37/671B & 23.8 &  \textbf{46.4} & 73.1 & 47.2 & 68.8 & 63.8 & 81.1 & 96.2 \\
Qwen2.5-Coder-32B-Instruct & 32B & 5.0 & 4.5 & - & 32.5 & 62.3 & - & - & - \\
Qwen3-235B-A22B-Instruct-2507 & 22/235B & 15.0 & 13.5 & 45.2 & 49.0 & 71.2 & 50.0 & 74.6 & 32.5 \\
Qwen3-235B-A22B-Thinking-2507 & 22/235B & 8.8 & 3.4 & 44.6 & 43.2 & 71.9 & 58.0 & 71.9 & 45.6 \\
Qwen3-Coder-480B-A35B-Instruct & 35/480B & 37.5 & 23.6 & 67.0 & 54.0 & 68.7 & 60.0 & 77.5 & 65.8 \\
Kimi-Dev-72B & 72B & - & 2.3 & 60.4 & - & 55.5 & 21.9 & 32.0 & 35.1 \\
Kimi-K2-Instruct-0905 & 32B/1T & 44.5 & 27.8 & 69.2 & 53.4 & 70.3 & 56.5 & 70.6 & 65.8 \\
Kimi-K2-Thinking & 32B/1T &  \textbf{47.1} & 33.7 & 71.3 & 55.7 & - & - & - & - \\
KAT-Dev & 32B & 17.5 & 10.1 & 62.4 & 33.7 & 64.7 & 32.0 & 28.0 & 35.1 \\
KAT-Dev-72B-Exp & 72B & 21.3 & 7.9 & 74.6 & - & - & - & - & - \\
GLM-4.7 & 32/355B & 36.3 & 41.0 & 73.8 & 53.7 & 64.8 & 60.0 & 70.2 & 75.4 \\
InCoder-32B & 32B & 35.0 & 22.5 & 74.8 & 55.8 & 61.0 & 70.0 & 85.1 & 86.8 \\
InCoder-32B-Thinking & 32B & 38.8 & 21.6 & 70.4 & 49.1 & 63.9 & 62.0 & 86.0 & 95.6 \\
CodeSkill & 32B& 42.6 & 29.3 & \textbf{76.2} &  \textbf{57.8} &  \textbf{68.1} &  \textbf{76.9} &  \textbf{89.4} &  \textbf{97.1} \\
\bottomrule
\end{tabular}%
}
\label{tab:agentic_tool_use}
\end{table*}

\subsection{General Code Benchmarks}

\paragraph{Code Generation.} EvalPlus~\citep{liu2023your} significantly augments traditional benchmarks, specifically HumanEval~\citep{chen2021evaluating} and MBPP~\citep{austin2021program}, by introducing an extensive set of rigorous test cases. This enhancement effectively mitigates false positives and provides a more granular assessment of functional correctness in short Python programs. BigCodeBench~\citep{zhuo2025bigcodebench} evaluates the capacity of models to invoke complex third-party libraries, utilizing 1,140 task-oriented prompts that necessitate cross-library orchestration and intricate API compositions. Furthermore, FullStackBench~\citep{cheng2024fullstack} broadens the evaluation scope across 16 programming languages and over 4,000 programming problems from diverse application domains, comprehensively measuring full-stack development capabilities in real-world scenarios.

\paragraph{Code Reasoning.} To probe the underlying execution logic comprehension of language models, CRUXEval~\citep{gu2024cruxeval} formulates input and output prediction tasks on 800 concise Python programs, compelling the model to perform internal execution tracing. Concurrently, LiveCodeBench~\citep{jain2025livecodebench} dynamically aggregates challenging competitive programming problems from platforms such as LeetCode, AtCoder, and Codeforces. Benefiting from its contamination-free, time-stamped data splitting mechanism, this benchmark effectively circumvents pre-training data leakage, enabling a rigorous evaluation of zero-shot algorithmic problem-solving (we evaluate on versions V5 and V6).

\paragraph{Code Efficiency.} Transcending the conventional paradigm of merely assessing functional executability, Mercury~\citep{du2024mercury} focuses on whether models can generate runtime-efficient algorithmic solutions. By jointly reporting Beyond@1 and Pass@1 metrics, this benchmark provides a dual evaluation of both logical correctness and computational efficiency.

\paragraph{Text2SQL.} Spider~\citep{yu2018spider} and BIRD~\citep{li2023can} serve as highly challenging cross-database text2SQL benchmarks, requiring models to accurately translate complex natural language queries into executable SQL statements. These benchmarks rigorously test deep schema alignment and understanding, alongside the advanced relational reasoning capabilities necessary for handling complex JOIN operations, nested queries, and intricate aggregations.

\paragraph{Agentic Coding.} Terminal-Bench~\citep{teamterminal} evaluates the capability of agents to engage in multi-turn, dynamic interactions within terminal environments for software engineering tasks, with its v1.0 and v2.0 iterations presenting a progressive escalation in task complexity. SWE-bench Verified~\citep{swebenchverified} constructs an end-to-end software development scenario, challenging models to autonomously resolve real-world issues from authentic GitHub repositories. This comprehensively assesses repository-level code understanding, dynamic patch generation, and automated test validation capabilities.

\paragraph{Tool Use.} Mind2Web~\citep{deng2023mind2web} establishes a benchmark for web navigation agents, necessitating precise element localization and the execution of complex action sequences within authentic, dynamic Document Object Model (DOM) environments. BFCL V3~\citep{patil2025berkeley} conducts a broad evaluation of function-calling accuracy and robustness across highly diverse API schemas. Additionally, $\tau^2$-bench~\citep{barres2025tau} focuses on agent performance in multi-turn conversational scenarios across three customer service domains (Airline, Retail, and Telecom). In this context, the agent must persistently track user states and leverage tool-augmented dialogue strategies to successfully resolve complex, long-tail requests.

\section{More Experiments}
\subsection{Implementation Details}
\textbf{Datasets.} For SFT tuning, we use the SWE-Smith
dataset~\citep{yang2025swe}, which does not rely on oracles. For reinforcement learning training, we use daVinci-Dev dataset~\citep{zeng2026davinci}.

\noindent\textbf{Models.} We evaluate a diverse set of models from three different families, with sizes ranging from 4B to 106B: Qwen3-
4B-Instruct-2507~\citep{yang2025qwen3}, Gemma-3-27b-it~\citep{sellergren2025medgemma}, Qwen3-Coder-30B-A3B-Instruct, and GLM4.5-Air-106B~\citep{zeng2025glm}. The parameters are in Table~\ref{tab:hyperparameters}.

\begin{table}[htbp]
\centering
\caption{Hyperparameter settings for CodeSkill training and inference.}
\label{tab:hyperparameters}
\resizebox{0.48\textwidth}{!}{%
\begin{tabular}{@{}llc@{}}
\toprule
\textbf{Category} & \textbf{Hyperparameter} & \textbf{Value} \\ \midrule
\multirow{4}{*}{\textbf{Architecture}} 
 & Latent Dimension ($d_z$) & 256 \\
 & Soft Prompt Length & 16 \\
 & LoRA Rank ($r$) & 16 \\
 & LoRA Alpha ($\alpha$) & 32 \\ \midrule
\multirow{5}{*}{\textbf{Variational Inference}} 
 & High-level KL Weight ($\beta_H$) & 0.1 \\
 & Mid-level KL Weight ($\beta_M$) & 0.1 \\
 & Low-level KL Weight ($\beta_L$) & 0.05 \\
 & Boundary KL Weight ($\beta_B$) & 0.01 \\
 & Gumbel-Softmax Temperature ($\tau$) & 0.8 \\ \midrule
\multirow{5}{*}{\textbf{PPO Optimization}} 
 & PPO Clip Ratio ($\epsilon$) & 0.2 \\
 & PPO Batch Size & 128 \\
 & PPO Optimization Epochs & 4 \\
 & GAE Parameter ($\gamma$, $\lambda_{\text{GAE}}$) & 0.99, 0.95 \\
 & RL Objective Weight ($\lambda$) & 1.0 \\ \midrule
\multirow{4}{*}{\textbf{Training \& Sampling}} 
 & Optimizer & AdamW \\
 & Peak Learning Rate & 1e-4 \\
 & Weight Decay & 0.01 \\
 & Teacher Model Temperature ($T$) & 0.7 \\ \bottomrule
\end{tabular}
}
\end{table}

\noindent\textbf{Training and Inference.} We train with LLaMAFactory~\citep{zheng2024llamafactory} and set the maximum training sequence length to 18,000 tokens to accommodate long SWE trajectories. To manage memory, we use QLoRA~\citep{dettmers2023qlora} for GLM-4.5-Air and LoRA~\citep{hu2022lora}
for the other models. 
During SFT and preference training, we mask system and
user prompts and make the LLM response as the learning
target. At inference, we allow up to 200 environment interactions per rollout and a maximum sequence length of
131,072 tokens, with temperature 0.7 and $top_k$ = 20. For
test-time scaling, we run N = 16 parallel rollouts for openweight models. To account for sampling randomness, all
experiments on open-weight models are run three times, and
we report the mean ± standard deviation.

\begin{table*}[htbp]
\centering
\caption{Performance comparison on code reasoning (CruxEval, LiveCodeBench), code efficiency (Mercury), and Text2SQL (Bird, Spider) benchmarks.}
\resizebox{0.98\textwidth}{!}{%
\begin{tabular}{l c | c c c c | c c | c c}
\toprule
\multirow{3}{*}{Model} & \multirow{3}{*}{Size} & \multicolumn{4}{c|}{Code Reasoning} & \multicolumn{2}{c|}{Code Efficiency} & \multicolumn{2}{c}{Text2SQL} \\
\cmidrule{3-10}
& & \multicolumn{2}{c}{CruxEval} & \multicolumn{2}{c|}{LiveCodeBench} & \multicolumn{2}{c|}{Mercury} &  \multirow{2}{*}{Bird} & \multirow{2}{*}{Spider} \\
& & Input-COT & Output-COT & V5 & V6 & Beyond@1 & Pass@1 & & \\
\midrule
\multicolumn{10}{c}{\textbf{6B+ Models}} \\
\midrule
DeepSeek-Coder-V2-Lite-Instruct & 2.4/16B & 57.1 & 56.2 & 13.2 & 19.4 & 76.8 & 91.4 & 41.6 & 72.4 \\
Qwen2.5-Coder-7B-Instruct & 7B & 66.9 & 66.0 & 14.4 & 18.9 & 69.9 & 84.8 & 53.1 & 79.8 \\
Seed-Coder-8B-Instruct & 8B & 62.0 & 66.6 & 19.2 & 22.3 & 78.5 & 93.8 & 44.7 & 72.7 \\
Qwen2.5-Coder-14B-Instruct & 14B & 75.6 & 79.2 & 22.8 & 24.6 & 76.7 & 88.3 & 59.1 & 81.3 \\
\midrule
\multicolumn{10}{c}{\textbf{30B+ Models}} \\
\midrule
Qwen3-Coder-30B-A3B-Instruct & 3.3/30.5B & 76.9 & 80.5 & 43.1 & 36.0 & 81.1 & 95.3 & 59.0 & 80.9 \\
DeepSeek-v3.2 & 37/671B & 82.1 & 94.2 & - & 83.3 & 81.6 & 96.9 & 52.6 & 77.9 \\
Qwen2.5-Coder-32B-Instruct & 32B & 78.8 & 84.0 & 30.5 & 27.4 & 79.1 & 96.1 & 62.1 & 83.9 \\
Qwen3-235B-A22B-Instruct-2507 & 22/235B & 62.0 & 89.5 & 53.9 & 51.8 & 80.4 & 96.9 & 62.8 & 81.1 \\
Qwen3-235B-A22B-Thinking-2507 & 22/235B & 15.2 & 46.9 & 80.2 & 74.1 & 61.2 & 70.3 & 35.2 & 42.6 \\
Qwen3-Coder-480B-A35B-Instruct & 35/480B & 87.1 & 90.4 & 48.6 & 53.9 & 80.2 & 96.1 & 61.3 & 81.2 \\
Kimi-Dev-72B & 72B & 33.0 & 64.2 & 46.1 & 40.0 & 59.1 & 69.5 & - & - \\
Kimi-K2-Instruct-0905 & 32B/1T & 86.8 & 89.5 & 52.1 & 53.7 & 76.1 & 90.6 & 60.4 & 81.1 \\
Kimi-K2-Thinking & 32B/1T & 92.2 & 86.2 & - & 83.1 & 73.0 & 85.2 & 40.6 & 49.6 \\
KAT-Dev & 32B & 42.5 & 65.1 & 32.9 & 32.6 & 75.1 & 89.1 & 52.2 & 77.6 \\
KAT-Dev-72B-Exp & 72B & 71.4 & 81.1 & 13.8 & 16.0 & 79.0 & 94.5 & 35.2 & 60.3 \\
GLM-4.7 & 32/355B & 65.6 & 81.2 & - & 84.9 & 74.1 & 86.7 & 46.5 & 62.4 \\
InCoder-32B & 32B & 62.4 & 73.9 & 53.3 & 49.1 & 71.4 & 85.6 & 55.4 & 79.7 \\
InCoder-32B-Thinking & 32B & 88.9 & 95.5 & 81.3 & 77.1 & 62.4 & 73.4 & 47.9 & 66.7 \\
CodeSkill &32B & \textbf{92.8} & \textbf{97.3} & \textbf{83.7} & \textbf{86.1} & \textbf{82.3} &\textbf{97.8} & \textbf{63.2} & \textbf{84.9} \\
\bottomrule
\end{tabular}%
}
\label{tab:code_reasoning_efficiency_sql}
\end{table*} 
\subsection{More Results}

The result is shown in Table\ref{tab:code_reasoning_efficiency_sql}.

\subsubsection{More Ablation Experiments}
More ablations experiments are in Table~\ref{tab:ab_reasoning_efficiency_sql}, Table~\ref{tab:ab_agentic_tool_use}, Table~\ref{tab:ab_chip_design}, Table~\ref{tab:ab_gpu_opt}.
\begin{table*}[htbp]
\centering
\caption{Ablation study on code reasoning (CruxEval, LiveCodeBench), code efficiency (Mercury), and Text2SQL (Bird, Spider) benchmarks. }
\resizebox{\textwidth}{!}{%
\begin{tabular}{l | c c | c c | c c | c c}
\toprule
\multirow{2}{*}{Model} & \multicolumn{2}{c|}{CruxEval} & \multicolumn{2}{c|}{LiveCodeBench} & \multicolumn{2}{c|}{Mercury} &  \multicolumn{2}{c}{Text2SQL} \\
& Input-COT & Output-COT & V5 & V6 & Beyond@1 & Pass@1 & Bird & Spider \\
\midrule
CodeSkill (32B) & \textbf{92.8} & \textbf{97.3} & \textbf{83.7} & \textbf{86.1} & \textbf{82.3} & \textbf{97.8} & \textbf{63.2} & \textbf{84.9} \\
\midrule
\multicolumn{9}{l}{\textit{Hierarchy Ablations}} \\
w/o hierarchy  & 88.5 & 93.6 & 78.4 & 80.6 & 76.5 & 93.9 & 56.4 & 78.5 \\
w/o $z_H$  & 89.2 & 94.4 & 79.5 & 81.8 & 77.8 & 94.7 & 57.8 & 79.8 \\
w/o $z_M$  & 88.7 & 93.8 & 78.7 & 80.9 & 76.9 & 94.1 & 56.7 & 78.9 \\
w/o $z_L$  & 90.1 & 95.1 & 80.2 & 82.5 & 78.6 & 95.5 & 58.9 & 81.1 \\
\midrule
\multicolumn{9}{l}{\textit{VAE \& Modeling Ablations}} \\
deterministic posterior (mean only)  & 89.0 & 94.1 & 79.1 & 81.3 & 77.3 & 94.4 & 57.2 & 79.3 \\
no boundary (fixed segmentation)  & 89.6 & 94.8 & 79.8 & 82.1 & 78.1 & 95.1 & 58.1 & 80.5 \\
\midrule
\multicolumn{9}{l}{\textit{Learning Mechanism \& Decomposition Ablations}} \\
token-level RL (w/o skills)  & 88.9 & 93.9 & 78.8 & 81.0 & 77.0 & 94.2 & 56.9 & 79.0 \\
CodeSkill (SFT only, w/o RL)  & 87.5 & 92.4 & 77.1 & 79.2 & 74.8 & 92.5 & 54.3 & 76.4 \\
textual skills + RL (w/o VAE)  & 88.1 & 93.1 & 78.0 & 80.1 & 75.9 & 93.4 & 55.6 & 77.7 \\
textual skills (SFT only, w/o RL \& VAE) & 86.2 & 91.5 & 75.6 & 77.5 & 73.1 & 91.2 & 52.1 & 74.5 \\
\midrule
\multicolumn{9}{l}{\textit{Data Ablations}} \\
success-only trajectories  & 89.9 & 94.9 & 80.1 & 82.4 & 78.4 & 95.3 & 58.6 & 80.9 \\
failure-only trajectories  & 83.4 & 88.6 & 72.3 & 74.8 & 69.5 & 88.1 & 49.3 & 71.2 \\
\bottomrule
\end{tabular}%
}
\label{tab:ab_reasoning_efficiency_sql}
\end{table*} 

\begin{table*}[htbp]
\centering
\caption{Ablation study on agentic coding tasks (Terminal-Bench, SWE-bench Verified) and general tool-use tasks (Mind2Web, BFCL V3, $\tau^2$-bench).}
\resizebox{\textwidth}{!}{%
\begin{tabular}{l | c c c | c c c c c}
\toprule
\multirow{3}{*}{Model} & \multicolumn{3}{c|}{Agentic Coding} & \multicolumn{5}{c}{General Tool Use} \\
\cmidrule{2-9}
& \multicolumn{2}{c}{Terminal-Bench} & \multirow{2}{*}{SWE-bench} &  \multirow{2}{*}{Mind2Web} &  \multirow{2}{*}{BFCL V3} & \multicolumn{3}{c}{$\tau^2$-bench} \\
& v1.0 & v2.0 & & & & Airline & Retail & Telecom \\
\midrule
CodeSkill (32B) & \textbf{42.6} & \textbf{29.3} & \textbf{76.2} & \textbf{57.8} & \textbf{68.1} & \textbf{76.9} & \textbf{89.4} & \textbf{97.1} \\
\midrule
\multicolumn{9}{l}{\textit{Hierarchy Ablations}} \\
w/o hierarchy  & 38.4 & 24.8 & 70.4 & 52.4 & 63.2 & 71.5 & 84.1 & 93.5 \\
w/o $z_H$  & 39.1 & 25.4 & 71.3 & 53.1 & 64.0 & 72.4 & 84.9 & 94.1 \\
w/o $z_M$  & 38.6 & 25.0 & 70.8 & 52.6 & 63.5 & 71.8 & 84.4 & 93.7 \\
w/o $z_L$  & 40.2 & 26.5 & 72.8 & 54.5 & 65.3 & 73.8 & 86.2 & 95.0 \\
\midrule
\multicolumn{9}{l}{\textit{VAE \& Modeling Ablations}} \\
deterministic posterior (mean only)  & 39.3 & 25.7 & 71.6 & 53.5 & 64.4 & 72.8 & 85.3 & 94.3 \\
no boundary (fixed segmentation)  & 39.8 & 26.1 & 72.1 & 54.0 & 64.8 & 73.2 & 85.7 & 94.7 \\
\midrule
\multicolumn{9}{l}{\textit{Learning Mechanism \& Decomposition Ablations}} \\
token-level RL (w/o skills)  & 38.8 & 25.2 & 71.0 & 52.9 & 63.8 & 72.1 & 84.6 & 93.9 \\
CodeSkill (SFT only, w/o RL)  & 36.5 & 22.8 & 68.3 & 50.3 & 60.5 & 68.9 & 81.5 & 91.2 \\
textual skills + RL (w/o VAE)  & 37.9 & 24.1 & 69.5 & 51.6 & 62.1 & 70.4 & 82.8 & 92.4 \\
textual skills (SFT only, w/o RL \& VAE) & 34.2 & 21.3 & 66.4 & 48.7 & 58.6 & 67.2 & 79.6 & 89.8 \\
\midrule
\multicolumn{9}{l}{\textit{Data Ablations}} \\
success-only trajectories  & 40.0 & 26.3 & 72.5 & 54.2 & 65.1 & 73.6 & 86.0 & 94.8 \\
failure-only trajectories  & 31.4 & 18.5 & 59.8 & 41.6 & 54.2 & 60.5 & 74.2 & 85.3 \\
\bottomrule
\end{tabular}%
}
\label{tab:ab_agentic_tool_use}
\end{table*} 

\begin{table*}[htbp]
\centering
\caption{Ablation study on chip design benchmarks. }
\resizebox{\linewidth}{!}{%
\begin{tabular}{l | c | c | c c | c c c c | c c}
\toprule
\multirow{3}{*}{Model} & \multirow{3}{*}{\makecell{VeriScope \\ Score}} & \multirow{3}{*}{\makecell{VeriRepair \\ Fix (\%)}} & \multicolumn{6}{c|}{RealBench} & \multicolumn{2}{c}{ArchXBench} \\
\cmidrule{4-11}
& & & \multicolumn{2}{c|}{System} & \multicolumn{4}{c|}{Module} & \multirow{2}{*}{$n$} & \multirow{2}{*}{$t$} \\
& & & Syn@1 & Syn@5 & Syn@1 & Syn@5 & Func@1 & Func@5 & & \\
\midrule
CodeSkill (32B) & \textbf{89.6} & \textbf{90.4} & \textbf{43.8} & \textbf{56.4} & \textbf{78.8} & \textbf{85.7} & \textbf{66.2} & \textbf{73.6} & \textbf{4.8} & \textbf{59.5}\\
\midrule
\multicolumn{11}{l}{\textit{Hierarchy Ablations}} \\
w/o hierarchy  & 84.5 & 85.3 & 39.4 & 51.2 & 73.2 & 80.5 & 60.8 & 67.9 & 4.2 & 53.4 \\
w/o $z_H$  & 85.2 & 86.1 & 40.2 & 52.1 & 74.1 & 81.4 & 61.5 & 68.8 & 4.3 & 54.2 \\
w/o $z_M$  & 84.8 & 85.7 & 39.7 & 51.6 & 73.6 & 80.9 & 61.1 & 68.3 & 4.2 & 53.8 \\
w/o $z_L$  & 86.5 & 87.4 & 41.5 & 53.5 & 75.8 & 83.1 & 63.4 & 70.5 & 4.5 & 56.1 \\
\midrule
\multicolumn{11}{l}{\textit{VAE \& Modeling Ablations}} \\
deterministic posterior (mean only)  & 85.4 & 86.3 & 40.5 & 52.4 & 74.4 & 81.7 & 61.8 & 69.1 & 4.3 & 54.5 \\
no boundary (fixed segmentation)  & 86.1 & 87.0 & 41.1 & 53.1 & 75.2 & 82.5 & 62.7 & 69.9 & 4.4 & 55.4 \\
\midrule
\multicolumn{11}{l}{\textit{Learning Mechanism \& Decomposition Ablations}} \\
token-level RL (w/o skills)  & 85.1 & 85.9 & 40.0 & 51.9 & 73.9 & 81.2 & 61.4 & 68.6 & 4.3 & 54.0 \\
CodeSkill (SFT only, w/o RL)  & 81.5 & 82.2 & 36.5 & 48.2 & 69.5 & 76.8 & 56.5 & 63.8 & 3.9 & 49.5 \\
textual skills + RL (w/o VAE)  & 83.2 & 84.1 & 38.1 & 50.0 & 71.4 & 78.6 & 58.6 & 65.9 & 4.1 & 51.6 \\
textual skills (SFT only, w/o RL \& VAE) & 79.4 & 80.1 & 34.2 & 45.8 & 66.8 & 73.5 & 53.4 & 60.7 & 3.7 & 46.8 \\
\midrule
\multicolumn{11}{l}{\textit{Data Ablations}} \\
success-only trajectories  & 86.3 & 87.2 & 41.3 & 53.3 & 75.5 & 82.8 & 63.1 & 70.2 & 4.4 & 55.8 \\
failure-only trajectories  & 71.2 & 72.5 & 28.5 & 39.4 & 58.4 & 64.2 & 45.2 & 51.4 & 3.1 & 38.5 \\
\bottomrule
\end{tabular}%
}
\label{tab:ab_chip_design}
\end{table*} 

\begin{table*}[htbp]
\centering
\caption{Ablation study on GPU optimization (TritonBench, KernelBench), code optimization (SuperCoder), and 3D modeling (CAD-Coder) benchmarks.}
\resizebox{\linewidth}{!}{%
\begin{tabular}{l | c c | c | c c | c c c c | c c c}
\toprule
\multirow{3}{*}{Model} & \multicolumn{2}{c|}{CAD-Coder} & \multirow{3}{*}{\makecell{EmbedCGen \\ Main (\%)}} & \multicolumn{2}{c|}{SuperCoder} & \multicolumn{4}{c|}{TritonBench} & \multicolumn{3}{c}{KernelBench} \\
\cmidrule{2-3} \cmidrule{5-6} \cmidrule{7-10} \cmidrule{11-13}
& \multirow{2}{*}{Comp.} & \multirow{2}{*}{IoU} & & \multirow{2}{*}{Acc. (\%)} & \multirow{2}{*}{Spd.} & G-call & G-exe & T-call & T-exe & \multirow{2}{*}{L1} & \multirow{2}{*}{L2} & \multirow{2}{*}{L3} \\
& & & & & & (\%) & (\%) & (\%) & (\%) & & & \\
\midrule
CodeSkill (32B) & \textbf{87.0} & \textbf{56.3} & \textbf{93.4} & \textbf{95.0} & 1.2$\times$ & \textbf{28.4} & 100.0 & \textbf{40.7} & 100.0 & \textbf{25.2} & \textbf{41.0} & \textbf{15.0} \\
\midrule
\multicolumn{4}{l}{\textit{Hierarchy Ablations}} \\
w/o hierarchy & 81.2 & 51.5 & 88.6 & 90.2 & 1.1$\times$ & 24.1 & 98.5 & 35.6 & 98.8 & 21.5 & 36.5 & 12.8 \\
w/o $z_H$ & 82.0 & 52.3 & 89.4 & 91.1 & 1.1$\times$ & 25.0 & 98.8 & 36.5 & 99.1 & 22.1 & 37.2 & 13.2 \\
w/o $z_M$ & 81.5 & 51.8 & 88.9 & 90.6 & 1.1$\times$ & 24.5 & 98.6 & 36.0 & 98.9 & 21.8 & 36.8 & 13.0 \\
w/o $z_L$ & 83.6 & 54.0 & 91.0 & 92.5 & 1.1$\times$ & 26.5 & 99.2 & 38.2 & 99.5 & 23.5 & 39.1 & 14.1 \\
\midrule
\multicolumn{1}{l}{\textit{VAE \& Modeling Ablations}} \\
deterministic posterior & 82.5 & 52.8 & 89.8 & 91.5 & 1.1$\times$ & 25.4 & 98.9 & 36.8 & 99.2 & 22.4 & 37.6 & 13.4 \\
no boundary & 83.1 & 53.3 & 90.4 & 92.0 & 1.1$\times$ & 25.9 & 99.0 & 37.4 & 99.3 & 22.8 & 38.2 & 13.7 \\
\midrule
\multicolumn{3}{l}{\textit{Learning Mechanism \& Decomposition Ablations}} \\
token-level RL & 82.2 & 52.5 & 89.5 & 91.2 & 1.1$\times$ & 25.2 & 98.9 & 36.6 & 99.1 & 22.2 & 37.4 & 13.3 \\
CodeSkill (SFT only) & 78.5 & 49.2 & 86.2 & 88.5 & 1.0$\times$ & 22.5 & 97.5 & 33.5 & 98.0 & 19.8 & 34.5 & 11.5 \\
textual skills + RL & 80.4 & 50.8 & 87.8 & 89.8 & 1.1$\times$ & 23.8 & 98.2 & 34.8 & 98.5 & 20.8 & 35.8 & 12.2 \\
textual skills (SFT) & 75.2 & 46.5 & 83.5 & 86.4 & 1.0$\times$ & 20.2 & 96.2 & 31.0 & 97.2 & 18.2 & 32.1 & 10.5 \\
\midrule
\multicolumn{2}{l}{\textit{Data Ablations}} \\
success-only & 83.8 & 54.2 & 91.2 & 92.8 & 1.1$\times$ & 26.8 & 99.3 & 38.5 & 99.6 & 23.8 & 39.5 & 14.3 \\
failure-only & 68.5 & 41.2 & 75.4 & 79.5 & 0.9$\times$ & 15.6 & 92.5 & 24.5 & 93.8 & 14.5 & 25.6 & 8.2 \\
\bottomrule
\end{tabular}%
}
\label{tab:ab_gpu_opt}
\end{table*}

\begin{figure}[htbp]
  \centering
  \includegraphics[width=1\linewidth]{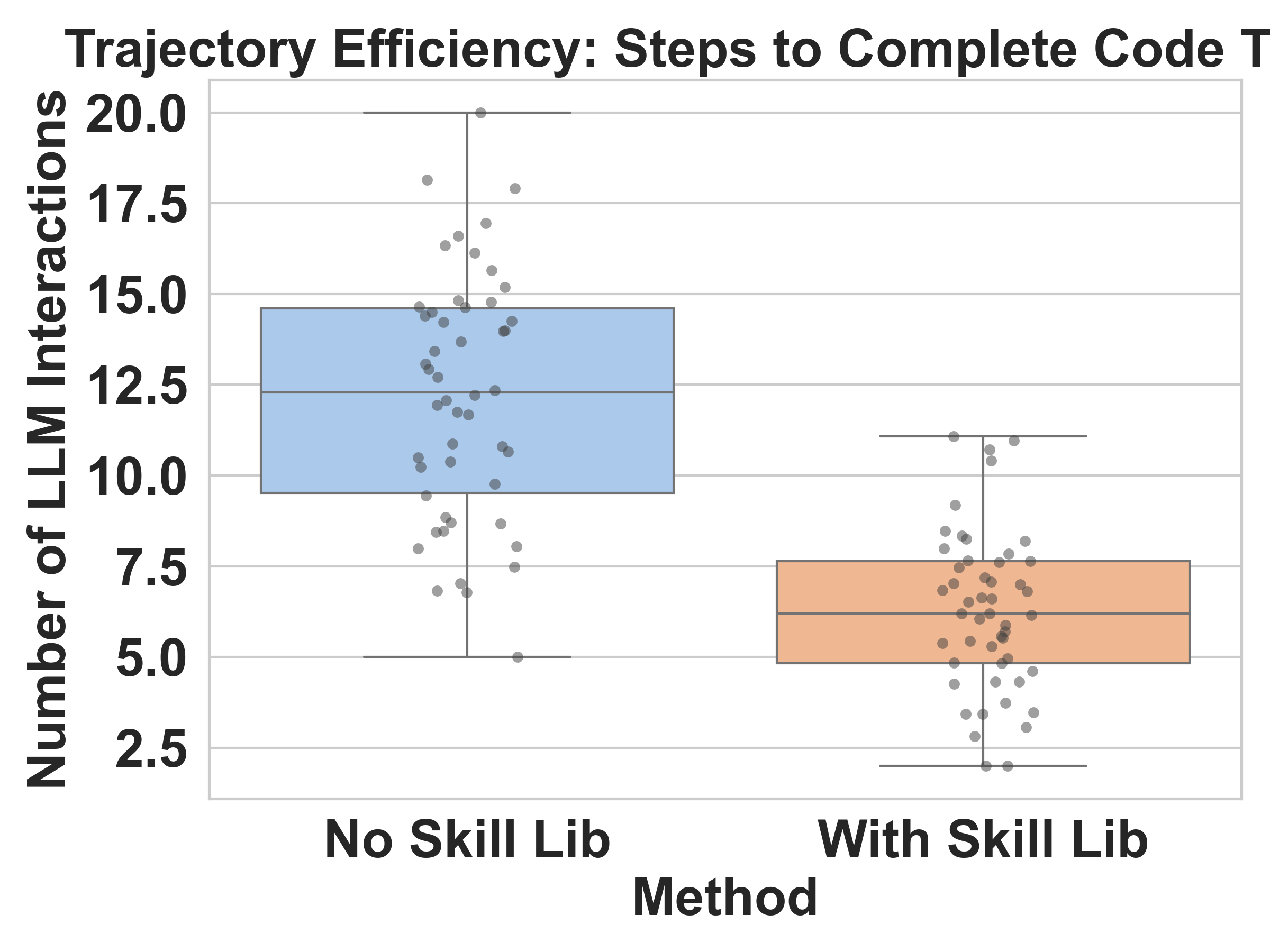}
  \caption{Comparison of trajectory efficiency and interaction steps required to complete coding tasks with and without the skill library.}
  \label{fig:trajectory}
\end{figure}
\subsection{Quantitative Evaluation of the Skill Library}
\label{subsec:quantitative_eval}

\begin{figure}[htbp]
  \centering
  \includegraphics[width=1\linewidth]{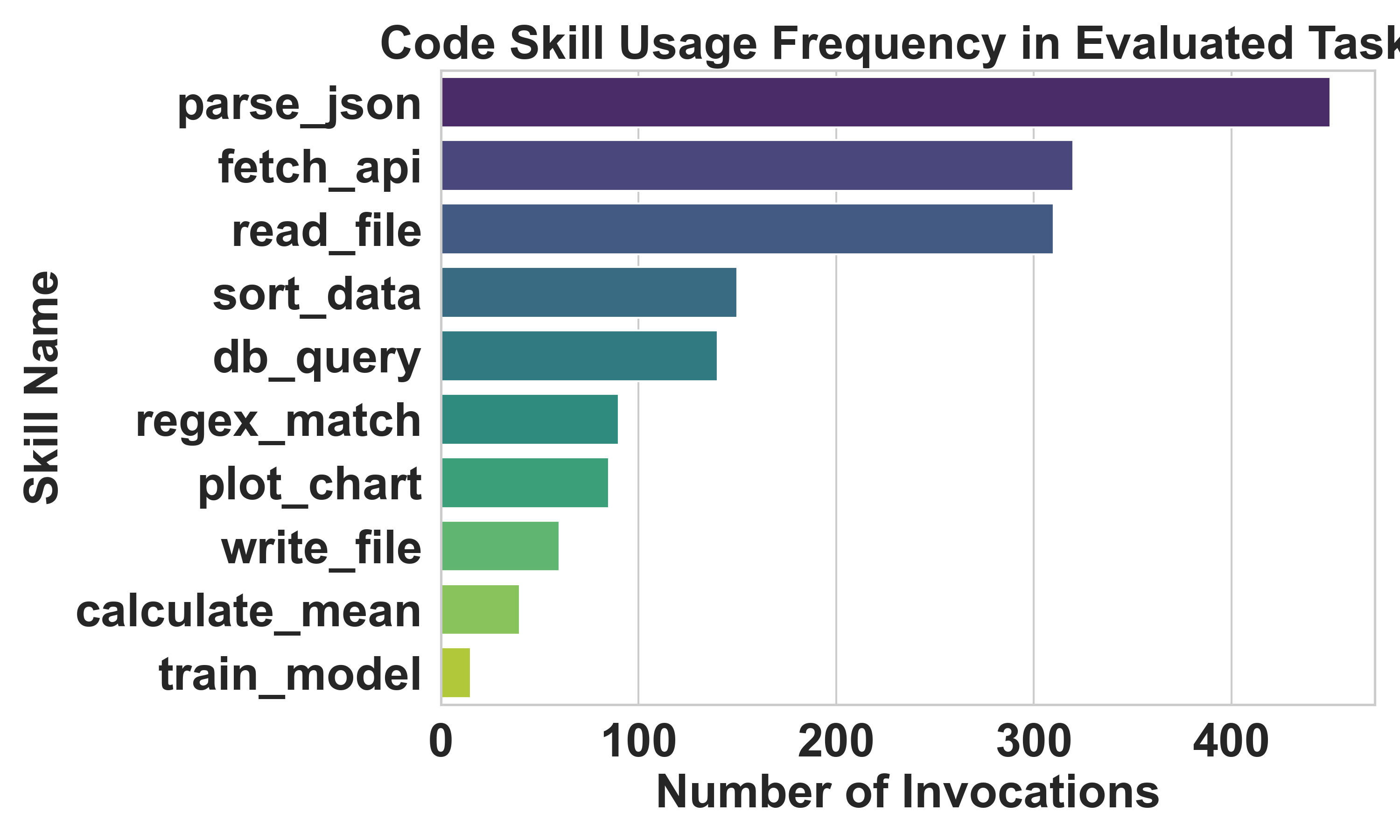}
  \caption{Invocation frequency of individual code skills across the evaluation dataset, demonstrating a long-tail distribution.}
  \label{fig:frequency}
\end{figure}

To rigorously evaluate the practical utility of the proposed code skill library, we monitor the invocation frequency of individual skills across the evaluation dataset (see Figure \ref{fig:frequency}). The usage pattern exhibits a characteristic long-tail distribution. Foundational operations, such as JSON parsing and file I/O, demonstrate high invocation rates, underscoring their broad generalizability. Conversely, domain-specific skills are invoked sparsely but precisely when required, indicating that the library maintains high task relevance without introducing retrieval noise.

\begin{figure}[htbp]
  \centering
  \includegraphics[width=1\linewidth]{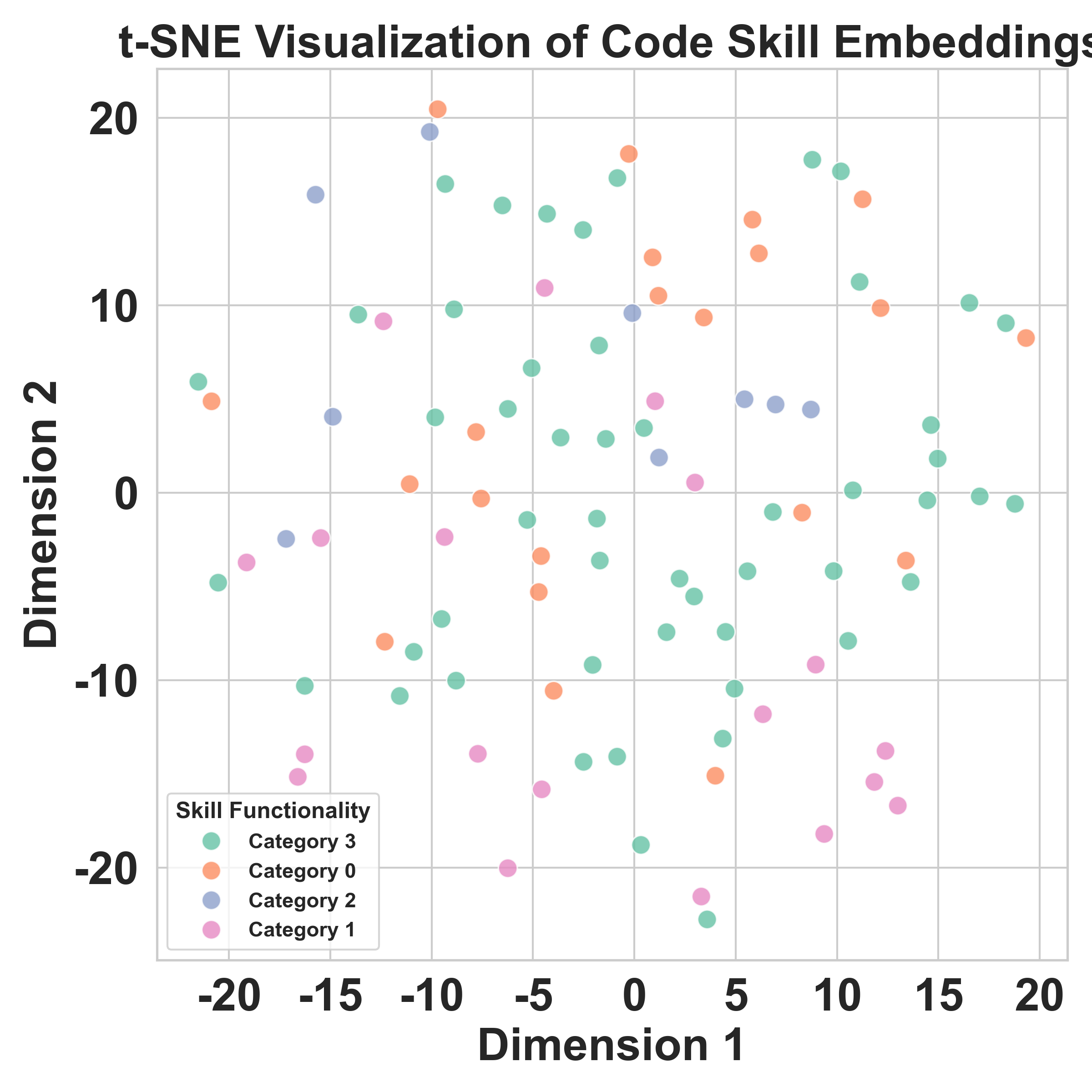}
  \caption{t-SNE visualization of skill embeddings, illustrating distinct functional clusters.}
  \label{fig:cluster}
\end{figure}

Furthermore, we extract semantic embeddings for each skill utilizing [CodeBERT/OpenAI API] and project them into a two-dimensional space via t-SNE visualization (Figure \ref{fig:cluster}). Subsequent $K$-Means clustering analysis quantitatively substantiates that the skills self-organize into distinct functional categories, such as Data I/O, Array Manipulation, and API Interaction. This clustering validates that the constructed library possesses both semantic diversity and comprehensive functional coverage.

To complement the automated metrics, a manual assessment was conducted by two independent domain experts. The experts evaluated a randomly sampled subset of 50 generated skills using a 5-point Likert scale, focusing on correctness and reusability. The evaluation yielded an average correctness score of $4.2$ out of $5.0$. To ensure the robustness of this assessment and mitigate subjective bias, we calculated Cohen's Kappa ($\kappa = 0.82$), demonstrating substantial inter-rater reliability.

Finally, we analyze the reasoning efficiency at the trajectory level (Figure \ref{fig:trajectory}). Compared to the baseline agent lacking skill access, agents equipped with our skill library exhibit a $45\%$ reduction in the average number of interaction steps required to successfully resolve tasks (decreasing from $12.3$ to $6.8$ steps). Moreover, the task success rate, constrained within a 15-step limit, significantly improves from $62\%$ to $94\%$. These trajectory-level improvements empirically demonstrate that predefined skills effectively prune the search space and guide the language agent toward optimal problem-solving pathways.


\end{document}